\documentclass[11pt]{article}

\usepackage[final]{acl}

\usepackage{times}
\usepackage{latexsym}
\usepackage{xspace}
\usepackage{xcolor}
\usepackage{booktabs}
\usepackage{tabularx}
\usepackage{graphicx}
\usepackage{amsmath}
\usepackage{amssymb}
\usepackage{pifont}
\usepackage{multirow}
\usepackage{array}
\usepackage{makecell}
\usepackage{enumitem}
\usepackage[table]{xcolor}
\usepackage{mdframed}
\usepackage{verbatim}
\usepackage[most]{tcolorbox}
\usepackage[export]{adjustbox}
\usepackage{listings}
\tcbuselibrary{listings}
\usepackage{anyfontsize}
\usepackage{tabularray}
\usepackage{pifont}
\usepackage{fontawesome5}

\definecolor{softred}{RGB}{252, 232, 232}
\definecolor{softgreen}{RGB}{232, 245, 233}
\newcommand{\cmark}{\ding{51}} 
\newcommand{\xmark}{\ding{55}} 

\usepackage[T1]{fontenc}

\usepackage[utf8]{inputenc}

\usepackage{microtype}

\usepackage{inconsolata}

\usepackage{graphicx}

\title{EDU-CIRCUIT-HW: Evaluating Multimodal Large Language Models on Real-World University-Level STEM Student Handwritten Solutions}

\author{
  \textbf{Weiyu Sun\textsuperscript{1}},
  \textbf{Liangliang Chen\textsuperscript{1}},
  \textbf{Yongnuo Cai\textsuperscript{1}},
  \textbf{Huiru Xie\textsuperscript{1}}, \\
  \textbf{Yi Zeng\textsuperscript{2}},
  \textbf{Ying Zhang\textsuperscript{1}}
  \\
  \\
  \textsuperscript{1}Georgia Institute of Technology,
  \textsuperscript{2}Virginia Tech
  \\
  \texttt{\{wsun355, liangliang.chen, ycai350, hxie77, yzhang\}@gatech.edu} \\
  \texttt{yizeng@vt.edu}
}

\newtcolorbox{solutionbox}[1]{
    enhanced,
    colback=gray!10,             
    colframe=gray!60!black,      
    title={#1},
    fonttitle=\bfseries,         
    top=2pt, bottom=2pt, left=2pt, right=2pt 
}

\newtcblisting{promptbox}[1]{
    colback=gray!10,
    colframe=gray!60!black,
    title=#1,
    fonttitle=\bfseries,
    enhanced,
    listing only,
    listing options={
        mathescape=false,
        basicstyle=\fontsize{8pt}{9pt}\selectfont\ttfamily, 
        breaklines=true,
        columns=fullflexible,
        keepspaces=true,
        escapeinside={(*}{*)}, 
        lineskip=-0.5pt, 
        literate={Ω}{{\textOmega}}1 {é}{{\'e}}1 {°}{{$^\circ$}}1 {±}{{$\pm$}}1
    },
    top=2pt, bottom=2pt, left=2pt, right=2pt 
}

\newtcblisting{promptboxmathfree}[1]{
    colback=gray!10,
    colframe=gray!60!black,
    title=#1,
    fonttitle=\bfseries,
    enhanced,
    listing only,
    listing options={
        mathescape=false,
        basicstyle=\fontsize{8pt}{9pt}\selectfont\ttfamily, 
        breaklines=true,
        columns=fullflexible,
        keepspaces=true,
        escapeinside={(*}{*)}, 
        lineskip=-0.5pt, 
        literate={Ω}{{\textOmega}}1 {é}{{\'e}}1 {°}{{$^\circ$}}1 {±}{{$\pm$}}1
    },
    top=2pt, bottom=2pt, left=2pt, right=2pt 
}

\newtcblisting{promptboxsmallfont}[1]{
    colback=gray!10,
    colframe=gray!60!black,
    title=#1,
    fonttitle=\bfseries,
    enhanced,
    listing only,
    listing options={
        basicstyle=\fontsize{6pt}{9pt}\selectfont\ttfamily, 
        breaklines=true,
        columns=fullflexible,
        keepspaces=true,
        showstringspaces=false,
        extendedchars=true,
        lineskip=-0.5pt, 
        literate={Ω}{{\textOmega}}1 {é}{{\'e}}1 {°}{{$^\circ$}}1 {±}{{$\pm$}}1
    },
    top=2pt, bottom=2pt, left=2pt, right=2pt 
}

\newtcblisting{contentbox}[1]{
    colback=gray!10,
    colframe=gray!60!black,
    title=#1,
    fonttitle=\bfseries,
    enhanced,
    breakable,             
    listing only,
    listing options={
        basicstyle=\fontsize{8pt}{9pt}\selectfont\ttfamily, 
        breaklines=true,
        columns=fullflexible,
        keepspaces=true,
        showstringspaces=false,
        extendedchars=true,
        mathescape=true,      
        lineskip=-0.5pt, 
        literate={Ω}{{\textOmega}}1 {é}{{\'e}}1 {°}{{$^\circ$}}1 {±}{{$\pm$}}1
    },
    top=2pt, bottom=2pt, left=2pt, right=2pt 
}

\newcommand{\dsname}{\texttt{EDU-CIRCUIT-HW}\xspace} 
\newcommand{\dscount}{1334\xspace} 
\newcommand{\dscounto}{513\xspace} 
\newcommand{\studentcounto}{11\xspace} 
\newcommand{\dscountt}{821\xspace} 
\newcommand{\studentcountt}{18\xspace} 

\newcommand{\myicon}[1]{\includegraphics[height=2ex]{#1}}

\begin{document}
\maketitle
\begin{abstract}
Multimodal Large Language Models (MLLMs) hold significant promise for revolutionizing traditional education and reducing teachers' workload. However, accurately interpreting unconstrained STEM student handwritten solutions with intertwined mathematical formulas, diagrams, and textual reasoning poses a significant challenge due to the lack of authentic and domain-specific benchmarks. Additionally, current evaluation paradigms predominantly rely on the outcomes of downstream tasks (e.g., auto-grading), which often probe only a subset of the recognized content, thereby failing to capture the MLLMs' understanding of complex handwritten logic as a whole. To bridge this gap, we release \dsname, a dataset consisting of 1,300+ authentic student handwritten solutions from a university-level STEM course. Utilizing the expert-verified verbatim transcriptions and grading reports of student solutions, we simultaneously evaluate various MLLMs’ upstream recognition fidelity and downstream auto-grading performance. Our evaluation uncovers an astonishing scale of latent failures within MLLM-recognized student handwritten content, highlighting the models' insufficient reliability for auto-grading and other understanding-oriented applications in high-stakes educational settings. As a potential solution, we present a case study demonstrating that leveraging identified error patterns to preemptively detect and correct recognition errors, while requiring only minimal human intervention (e.g., routing 3.3\% of assignments to human graders and the remainder to the GPT-5.1 grader), can effectively enhance the robustness of the deployed AI-enabled grading system.

\begin{center}
\begin{tabular}{cl}
    \faGlobe & \href{https://gt-learning-innovation.github.io/CIRCUIT_EDU_HW_ACL}{\texttt{Project Website}} \\[0.5em]
    \faGithub & \href{https://github.com/gt-learning-innovation/CIRCUIT_EDU_HW_ACL}{\texttt{GitHub Repository}}
\end{tabular}
\end{center}
\end{abstract}

\section{Introduction}

\begin{figure}[ht!]
    \centering
    \includegraphics[width=1\linewidth]{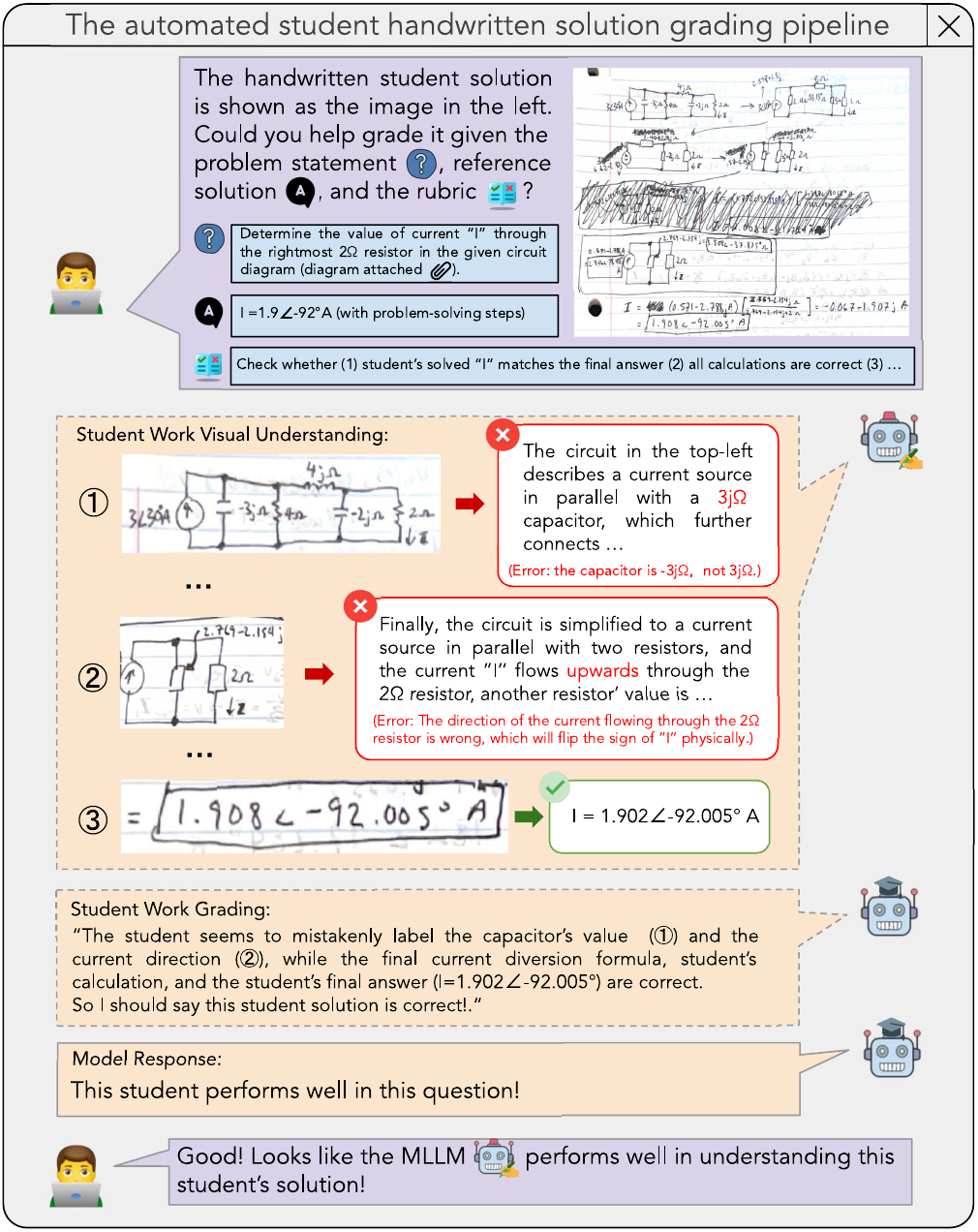}
    \caption{An exemplary STEM student handwritten solution auto-grading workflow: MLLM (e.g., \texttt{Gemini-2.5-Pro}) recognizes \ding{172}, \ding{173}, \ding{174}, and other visual contents of the solution, which serve as inputs for the consequent AI grading. Notably, recognition errors (\ding{172} and \ding{173}) may not influence the grading if they fall outside the specific grading criteria. As a result, this successful grading masks underlying recognition failures, leading users to overestimate the MLLM's visual understanding performance on the student solution.}
    \label{fig:intro_data_showcase}
\end{figure}

Recent cutting-edge multimodal large language models (MLLMs) have demonstrated near-human performance over a wide range of low-stakes, daily visual understanding tasks. However, as the focus shifts to high-stakes, precision-critical domains, the visual understanding of these MLLMs remains insufficiently robust and inadequately evaluated. One particularly challenging and underexplored setting is university-level STEM (Science, Technology, Engineering, and Mathematics) student handwritten solution understanding, where the model must interpret the complex, interleaved ``visual language'' consisting of unconstrained handwriting, intricate mathematical derivations, and hand-drawn diagrams. The interpreted content further serves as input to consequent downstream tasks such as auto-grading \citep{kortemeyer2024grading, khrulev2025check, chen2025benchmarking, lchen2025fie} as shown in Figure \ref{fig:intro_data_showcase}. To this end, a reliable and trustworthy MLLM-based workflow to understand student handwritten solutions is a longstanding ambition in AI-enabled education.

Despite the potential of MLLMs, evaluating their effectiveness in real-world university settings faces two critical hurdles. First is the scarcity of proper authentic data. Most existing benchmarks fall short of capturing the complexity of real-world student handwritten work. For instance, many benchmarks \citep{2023crohme, MathWriting} focus on isolated visual elements (e.g., a single mathematical expression in Figure \ref{fig:intro_data_showcase} \ding{174}) rather than the tightly interleaved reasoning processes characteristic of authentic student works (e.g., the entire input student solution in Figure \ref{fig:intro_data_showcase}). In addition, these benchmarks often omit or oversimplify hand-drawn diagrammatic understanding (e.g., \ding{172} and \ding{173} in Figure \ref{fig:intro_data_showcase}), despite diagrams playing a critical role in many STEM domains. Moreover, many prior benchmarks \citep{barallucy2024drawedumath} target relatively low-difficulty settings such as K–12-level math. As a result, they fail to reflect the challenges posed by college STEM handwritten assignments. Consequently, such benchmarks provide limited insight into the true visual understanding and reasoning capabilities of MLLMs under realistic, high-stakes educational conditions.

Second, there is a fundamental misalignment in evaluation paradigms. Most existing benchmarks for student handwritten solution understanding focus on a single downstream objective (e.g., VQA \citep{barallucy2024drawedumath} or auto-grading). While these task-specific evaluations are valuable, such downstream tasks often probe only a subset of the recognized content, leaving many recognition errors unobserved if they do not affect the specific task outcome. One such example is shown in Figure \ref{fig:intro_data_showcase}, where recognition errors in \ding{172} and \ding{173} are not reflected in the final grading report due to the rubric limitations. Crucially, these ``negligible'' errors in the auto-grading task may be catastrophic for other downstream tasks, such as circuit-to-netlist transformation \citep{xu2025image2net}.

To bridge the abovementioned gaps, we introduce \dsname, a diagnostic dataset grounded in real-world university STEM education, which comprises \dscount authentic student handwritten solutions to complex circuit analysis problems. To enable fine-grained MLLM visual recognition ability analysis, we curated near-verbatim transcriptions for a data subset and utilize them as the reference to capture different MLLMs' errors in their recognized student solutions. Moreover, we build a taxonomy for these captured errors and evaluate their cascading impact on a downstream auto-grading task.

Our evaluation reveals a pervasive presence of latent recognition errors beneath current grading outcomes. Although these failures may remain dormant under coarse assessment criteria (e.g., binary correctness check), they become increasingly detrimental as grading granularity refines (e.g., specific, targeted score deduction), where higher precision inevitably exposes these underlying defects and degrades grading quality. This finding indicates that reliable student solution understanding and its applications remain a distant goal. In solution, we demonstrate the feasibility to detect and suppress unseen potential recognition failures via our identified error patterns in a case study.

Overall, our contributions are summarized as follows:

i) We release \dsname, a dataset of 1300+ authentic university-level STEM solutions, facilitating the evaluation of MLLMs' visual understanding capabilities in real-world educational scenarios.

ii) We conduct a comprehensive evaluation of various cutting-edge MLLMs' capabilities in handwriting understanding and downstream auto-grading on \dsname via a proposed automated diagnostic workflow. By establishing a fine-grained taxonomy of visual perception failures, we systematically analyze the cascading impact of recognition errors on the downstream auto-grading, which uncovers the latent risks concealed by seemingly robust downstream performance.

iii) We conduct a targeted case study to demonstrate that the identified error patterns can be leveraged to detect and prevent MLLMs' recognition failures over unseen student solutions, thereby enhancing the reliability of deployed grading systems.

\section{The \dsname Dataset}
\label{sec: dataset}
This section introduces the \dsname dataset for evaluating MLLMs' capabilities on university-level STEM handwritten solution recognition and downstream auto-grading tasks.

\subsection{Handwritten Solution Collection}
\label{sec: handwritten_solution_collection}
The \dsname dataset consists of \dscount handwritten homework solutions from 29 students in an undergraduate-level circuit analysis course at a large, public, research-intensive institution in the Southeast United States during the Spring 2025 semester. The homework problems are all from the textbook \citep{svoboda2013introduction} with the topics varying from the basic circuit concepts and elements to advanced topics such as the first- and second-order circuit analyses. Each handwritten solution figure in this dataset corresponds to one student's submission to a specific homework problem. In the pre-processing step, we removed the personally identifiable information of students, including students' names and university IDs, from the solution images. All the students' solutions are associated with expert-labeled reference grades in five different aspects, making the dataset able to be used to verify the performance of MLLM-enabled auto-graders, as illustrated in Section \ref{sec: auto-grading intro}. Since students may leverage matrix theory, calculus, complex operations, and hand-drawn diagrams when solving the circuit problem, the benchmarks and analyses built on the \dsname can also be extended to other related STEM areas beyond circuit analyses.

\subsection{Dataset Statistics}
There are two data groups in the \dsname dataset: observation set and test set. The observation set consists of \dscounto handwritten solutions from \studentcounto students. Each solution in the observation set is associated with detailed grades and image recognition results, of which both are verified and proofread manually by experts, as detailed in Section \ref{sec: method}. Particularly, the expert-proofread recognition contains a near-verbatim transcription of all student-handwritten content, including natural-language descriptions for any non-textual elements such as hand-drawn circuit diagrams with annotations or function graphs. Since the recognized content is manually verified, the observation set can be regarded and utilized as a ``training set'' from which we can summarize the recognition patterns and use them to further improve the performance of both recognition and automated grading, as illustrated in Sections \ref{sec: method}--\ref{sec: case_study}. In addition to the observation set, the \dscountt samples from the remaining 18 students constitute the test set, in which each handwritten solution is associated with a ground truth grade but not expert-verified recognition. Table \ref{tab:dataset_summary} summarizes some key attributes of these two data groups. 

\begin{table}[t]
\centering
\resizebox{1\columnwidth}{!}{%
\begin{tabular}{lcc}
\toprule
\textbf{Attribute} & \textbf{Observation Set} & \textbf{Test Set} \\
\midrule
Number of Unique Questions & 62 & 62 \\
Number of Students & \studentcounto & \studentcountt \\
Number of Samples & \dscounto & \dscountt \\
w/ Verified Recognition & \cmark & \xmark \\
w/ Ground Truth Grade & \cmark & \cmark \\
Student's Solution Accuracy & 72.12\% & 71.86\% \\
\bottomrule
\end{tabular}
}
\caption{Key attributes of the observation and test sets in \dsname.}
\label{tab:dataset_summary}
\end{table}

\section{Handwritten Solution Recognition and Automated Grading Framework}
\label{sec: method}
Section \ref{sec: dataset} describes the collection and basic statistics of the \dsname dataset. In this section, we first detail how the handwritten solution can be effectively recognized by prompting MLLMs strategically. We then describe how the recognition errors can be classified into four well-defined categories and propose a method to investigate their impact on the downstream auto-grading task. 

\subsection{Handwritten Solution Recognition}
\label{sec: Recognizing the Handwritten Content}
Reliable handwritten solution recognition is essential and further affects the downstream task, such as homework auto-grading. However, even the most cutting-edge MLLMs perform poorly on STEM handwritten homework solution recognition tasks \citep{barallucy2024drawedumath, caraeni2024evaluating}.

\begin{figure}[t]
    \centering
    \includegraphics[width=1\columnwidth]{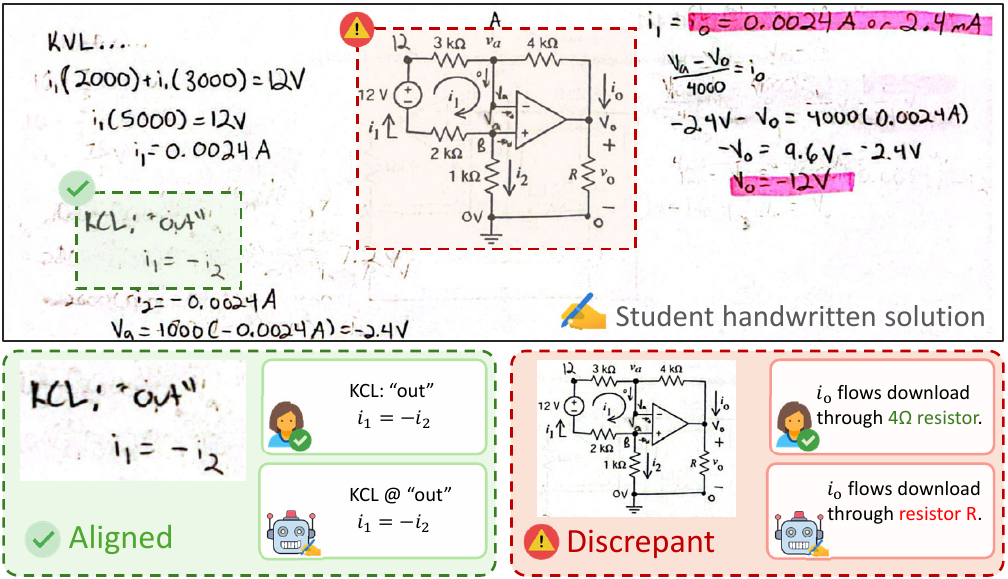}
    \caption{The demonstration on our recognition error detection. Given a handwritten solution, the \raisebox{-0.3ex}{\myicon{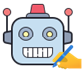}} MLLM's recognition result is compared with the \raisebox{-0.3ex}{\myicon{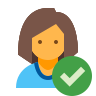}} expert-verified transcriptions in a full-text level, after which all discrepant items within the recognition result are listed as recognition errors. Note that we regard two items as aligned when they are semantically equivalent in engineering education (e.g., \raisebox{-0.3ex}{\myicon{images/icons8-confirm-96.png}} KCL: ``out'' and \raisebox{-0.3ex}{\myicon{images/MLLM_recognizer.png}} KCL: @ ``out'' are regarded as semantically aligned despite the additional ``@'').}
    \label{fig:model_difference}
\end{figure}

In this section, we analyze MLLMs' recognition performance in a more fine-grained view. To this end, after obtaining MLLMs' recognition results for the observation set, we build an LLM-as-a-judge pipeline to locate all potential recognition errors, which directly reflect the handwritten segments that the MLLM has trouble recognizing, as shown in Figure \ref{fig:model_difference}. Specifically, we first prompt \texttt{Gemini-2.5-Pro} to recognize the image-based students' solution submissions and output the resulting textual descriptions in the Markdown format.
In this process, the equations and texts are prompted to be recognized verbatim, and the diagrams, if existing, are interpreted in natural language that contains the full diagram information, such as the circuit topology and students' annotations. The recognition results are then manually proofread against the original handwritten images by experts who are also responsible for rectifying the identified recognition errors, such as wrong diagram depictions or equations, by saving the rectified version in new Markdown files.

The expert-proofread Markdown files serve as the ground truth to assist an LLM-judger, which we use \texttt{Gemini-2.5-Pro}, to capture the potential recognition errors from the evaluated MLLM. Specifically, for each handwritten solution, we provide the LLM-judger with both the verified recognition and the evaluated MLLM's recognition and ask the LLM-judger to list all potential discrepant recognition items between them. Note that an item in this scenario can be either a sentence or an equation. Some item examples can be found in Figure \ref{prompt: example_about_recognition_error} in Appendix \ref{experiment_details}.

\subsubsection{LLM-as-a-Judge Method Validation}
In order to verify the LLM-judger's reliability, we further manually check the recognition discrepancies and compare the identified discrepancies with those listed by the LLM-judger. The comparisons are made at two levels: sample level and item level. The sample level aims to measure the agreement between the LLM-judger and human experts in determining whether a handwritten recognition result contains any recognition errors. We use a binary indicator to describe this agreement. The item level focuses on whether each individual recognition error annotated by human experts (e.g., the discrepant case shown in Figure \ref{fig:model_difference}) can be detected by the LLM-as-a-judge method. At this level, we evaluate error detection performance using the precision, recall, and the F1 score, where true positives (TP) are recognition errors identified by both the human expert and the LLM-judger, false positives (FP) are errors reported by the LLM-judger but not annotated by the human experts, and false negatives (FN) are errors annotated by the human experts but missed by the automated LLM-judger.

The comparison is conducted with over 186 recognized handwritten solution samples containing more than 5000 items. Specifically, we sample one student's solution randomly for all the 62 problems in our dataset and prompt 3 commercial MLLMs (\texttt{GPT-5.1}, \texttt{Claude-4.5-Sonnet}, and \texttt{Qwen3-VL-PLUS}) to make recognitions. For each recognized result, an expert manually annotates all recognition errors (like those discrepant items in Figure \ref{fig:model_difference}). These annotations are then used to evaluate the performance of LLM-judger at both sample and item levels. The results are shown in Table \ref{tab:human_vs_automatic}, which indicates both high sample-level accuracy (larger than 0.95 on all three models) and item-level consistency (precision, recall, and the F1 score are all close to or larger than 0.9 on three models). A closer inspection also indicates that false positives and false negatives are predominantly associated with ambiguous items rather than reflecting systematic misjudgments. These results suggest that the LLM-as-a-judge pipeline is highly reliable and thus enables automated evaluation of handwritten recognition performance of different MLLMs with human-level consistency. 

\begin{table}[t]
    \centering
    \setlength{\tabcolsep}{4pt} 
    \resizebox{1\columnwidth}{!}{%
        \renewcommand{\arraystretch}{1.2}
        \begin{tabular}{l c cccccc}
            \toprule
            \multirow{2}{*}{\textbf{Model}} & \textbf{Sample} & \multicolumn{6}{c}{\textbf{Item-Level Metrics}} \\
            \cmidrule(lr){2-2} \cmidrule(lr){3-8}
             & \textbf{Accuracy} & \textbf{TP} & \textbf{FP} & \textbf{FN} & \textbf{Precision} & \textbf{Recall} & \textbf{F1 Score} \\
            \midrule
            GPT-5.1 & 0.9516 & 97 & 6 & 10 & 0.9417 & 0.9065 & 0.9238 \\
            Claude-4.5-Sonnet & 0.9839 & 137 & 13 & 15 & 0.9133 & 0.9013 & 0.9073 \\
            Qwen3-VL-Plus & 0.9516 & 62 & 7 & 5 & 0.8986 & 0.9254 & 0.9118 \\
            \midrule
            \textbf{Total} & \textbf{0.9624} & \textbf{296} & \textbf{26} & \textbf{30} & \textbf{0.9193} & \textbf{0.9080} & \textbf{0.9136} \\
            \bottomrule
        \end{tabular}
    }
    \caption{Validation results of the automated recognition error detection via LLM-judgers against expert annotations across 186 handwritten solutions from three MLLMs.}
    \label{tab:human_vs_automatic}
\end{table}

\subsubsection{Taxonomy on Recognition Errors} 
Recognition errors vary from trivial character mistakes to severe logical misunderstandings. To gain detailed insights, we further divide the recognition errors into four categories in Table \ref{tab:error_taxonomy}. An LLM is prompted with clear category definitions to categorize all discrepant items in an automated way. With this taxonomy, we can observe the distribution of errors among different recognitions, which reveals the weaknesses of current MLLMs.

\begin{table}[t]
\centering
\renewcommand{\tabularxcolumn}[1]{m{#1}}

\renewcommand{\arraystretch}{1.0}

\newcommand{\equalheightstrut}{\rule[-2.2ex]{0pt}{5.5ex}}

\scriptsize 
\begin{tabularx}{0.475\textwidth}{
    >{\centering\arraybackslash\hsize=0.275\hsize\bfseries}X 
    >{\raggedright\arraybackslash\hsize=0.925\hsize}X
}
\toprule
\textbf{Category} & \multicolumn{1}{c}{\textbf{Introduction \& Example}} \\ \midrule

\vspace{-0.1cm}\makecell{Symbolic \&\\ Character} & Misreading alphanumeric glyphs, operators, or units. Example: $20000 \to 2a000$; $-V \to V$ \\ \midrule

\makecell{Structural \&\\ Notational} & Wrong mathematical layout or variable consistency. Example: $\text{R} = \frac{\frac{1}{8}}{\frac{1}{8} + \frac{1}{16}} \Omega \to \text{R} = \frac{8}{8 + 16} \Omega$ \\ \midrule

Diagrammatic & Misinterpretation of diagram space attributes or annotations. Example: Erroneous circuit topology. \\ \midrule

\vspace{-0.1cm}\makecell{Textual \&\\ Logical} & Mislabeling context or sequence of reasoning. Example: Omission of critical derivation steps \\ \bottomrule
\end{tabularx}
\caption{Taxonomy of recognition errors. More examples with image illustrations are available in Tables \ref{tab:symbol_error_examples}--\ref{tab:diagrammatic_error_logical} in Appendix \ref{sec: example_for_error_taxonomy}.}
\label{tab:error_taxonomy}
\end{table}

\subsection{Downstream Task: Automated Grading}
\label{sec: auto-grading intro}
Section \ref{sec: Recognizing the Handwritten Content} described the methods of handwritten solution recognition and detection for recognition errors. In this section, we will explore how the upstream recognition errors affect the downstream task, which is helpful to minimize the negative effect of the recognition errors and boost the downstream task performance.


In this work, we consider auto-grading as the downstream task, a process widely implemented across educational institutions worldwide. Specifically, we use an LLM-based grader, which is provided with MLLM-recognized student solutions and the problem context to assess student performance on the given problem. In our settings, the problem context includes the problem description, reference solution, and grading rubrics. The auto-grader’s reports are then compared with expert-verified reference grades available in the dataset, as described in Section \ref{sec: handwritten_solution_collection}.

In the \dsname dataset, each student's handwritten solution was evaluated by a human expert based on a multi-dimensional rubric that covers five distinct perspectives of assessment as shown in Table \ref{tab:rubric}. With these five grading aspects, the rubrics are defined to be problem-specific by using a unique rubric for each problem in our dataset. Given the rubric, the human expert will deduct scores for those students' submissions violating the rubric per perspective. For example, ``\texttt{\{E:0.02pts, U:0.01pts\}}'' indicates that the student might use a wrong equation (e.g., incorrect Ohm's Law ``$R=\frac{I}{V}$'') and an incompatible unit (e.g., using ``voltage'' instead of ``ampere'' to describe a current). These expert-verified grading results can serve as ground truth to evaluate the performance of the auto-grading pipeline.

\begin{table}[t]
    \scriptsize 
    \centering    
    \resizebox{1\columnwidth}{!}{%
        \begin{tabular}{cc}
            \toprule
            \textbf{Type} & \textbf{Description} \\ 
            \midrule
            \textbf{E}   & Equation Error: Incorrect or inapplicable equations \\
            \textbf{M}   & Method Error: Flawed problem-solving logic \\
            \textbf{U}   & Unit Error: Missing or incorrect units \\
            \textbf{C}   & Calculation Error: Wrong arithmetic with the correct equation \\
            \textbf{NC}  & Incomplete Solution: Incomplete answer or missing steps \\
            \bottomrule
        \end{tabular}%
    }
    \caption{The rubric for grading students' work. A student's solution can violate multiple standards in the table and get scores deducted separately. A rubric example can be found in Figure \ref{prompt: rubric example} in Appendix \ref{experiment_details}.}
    \label{tab:rubric}
\end{table}

\begin{figure*}[t]
    \centering
    \includegraphics[width=1\linewidth]{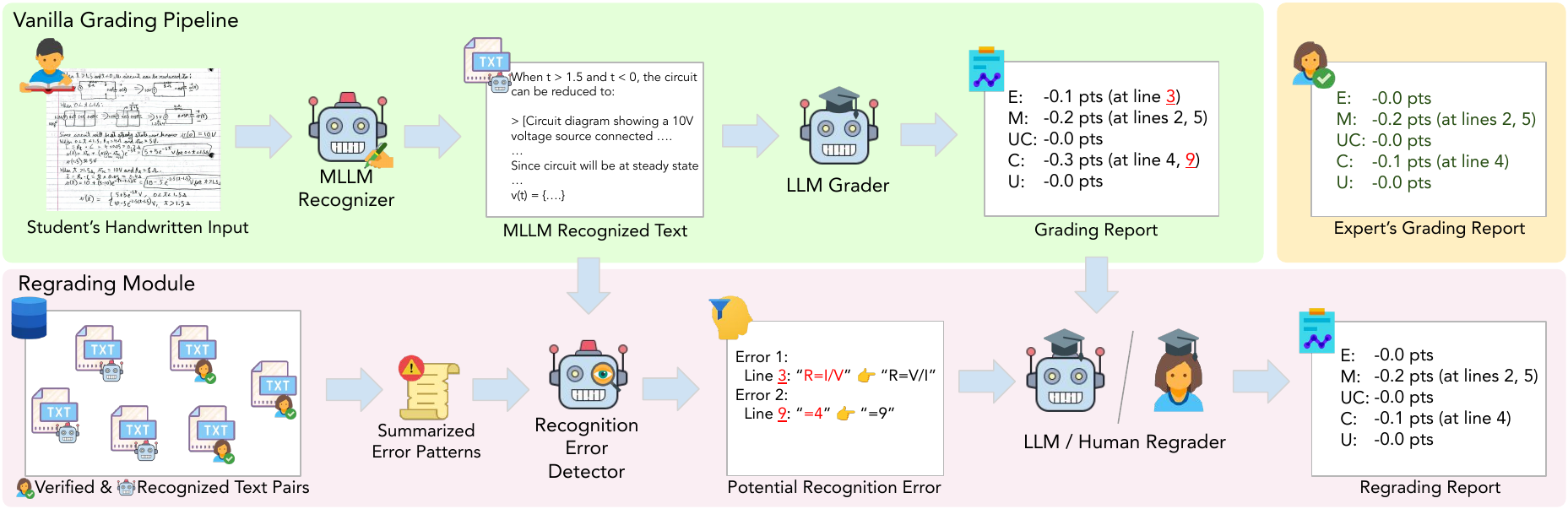} 
    \caption{The proposed auto-grading pipeline. The ``vanilla grading pipeline'' in green box is a widely-used auto-grading paradigm in AI-enabled education field. In our experiment (Section \ref{sec: exp_recognition}), we detect and analyze recognition errors within ``MLLM Recognized Text'' utilizing the automated pipeline introduced in Section \ref{sec: Recognizing the Handwritten Content}, and further evaluate ``Grading Report'' by benchmarking it against the ``Expert's Grading Report'' in yellow box. In the pink box, we present an exemplary regrading module in which human experts first summarize the detected recognition error patterns and then leverage them (as detailed in Figures \ref{prompt: recog err detection (Part I)} and \ref{prompt: recog err detection (Part II)}) to filter potential recognition errors within the ``MLLM Recognized Text.'' Solutions flagged as containing potential recognition errors are subsequently reassessed by LLM- or human-based regraders. Additional details are provided in Section \ref{sec: case_study}.}
    \label{fig:exp_overall_pipeline}
\end{figure*}

\begin{table*}[ht]
\centering
\label{tab:ocr_grading_comparison}
\resizebox{0.95\linewidth}{!}{%
\begin{tabular}{llcccccc}
\toprule
\multirow{2.4}{*}{\textbf{Grader}} &
\multirow{2.4}{*}{\textbf{Visual Recognizer}} &
\multicolumn{2}{c}{\textbf{Recognition Quality}} &
\multicolumn{3}{c}{\textbf{Grading Performance}} & 
\multirow{2.4}{*}{\textbf{EIR (\%)$\downarrow$}} \\ 
\cmidrule(lr){3-4} \cmidrule(lr){5-7} 
 &  & \textbf{SER (\%)$\downarrow$} & \textbf{AEC$\downarrow$} &
\textbf{Binary (\%)$\uparrow$} & \textbf{Type (\%)$\uparrow$} & \textbf{Point (\%)$\uparrow$} & \\ 
\midrule
\texttt{Graduate} & -- & -- & -- & 83.63 & \textbf{82.46} & \textbf{81.29} & -- \\

\midrule
\multirow{7}{*}{\texttt{GPT-5.1}}
 & \texttt{Human Expert} & -- & -- & \textbf{89.47} & 78.36 & 74.46 & -- \\
 & \texttt{Gemini-3-Preview} & \textbf{37.62} & \textbf{0.61} & 87.91 & 78.17 & 74.27 & 7.60\% \\
 & \texttt{Gemini-2.5-Pro} & 53.52 & 1.23 & 85.58 & 73.68 & 69.40 & 14.72\% \\
 & \texttt{GPT-5.1} & 71.54 & 2.05 & 77.78 & 65.50 & 61.99 & 17.89\% \\
 & \texttt{Qwen3-VL-Plus} & 61.72 & 1.38 & 80.90 & 68.62 & 65.11 & 16.67\% \\
 & \texttt{Claude-4.5-Sonnet} & 80.70 & 2.76 & 77.58 & 63.16 & 59.84 & 18.05\% \\
 & \texttt{Qwen3-VL-8B-Thinking} & 85.43 & 2.79 & 75.05 & 61.01 & 56.92 & 19.60\% \\
\bottomrule
\end{tabular}
}
\caption{Overall evaluation results of different MLLMs' recognition quality on student handwritten works in \dsname (the observation set), and their impact (i.e., Error Impact Rate (EIR)) on the downstream grading performance. We further add two baselines: (1) A \texttt{Graduate} student grades directly on student handwritten samples and (2) \texttt{Human Expert} recognizes and provides the ``gold standard'' recognized text for the LLM grader for reference. For metrics, we adopt Sample Error Rate (SER) and Average Error Count (AEC) to evaluate recognition quality, and binary, type, and point agreements to evaluate grading performance. The \textbf{bold} values are the best scores across different metrics. The detailed taxonomic EIR analysis over different error categories is detailed in Figure \ref{fig:error distribution}.}
\label{tab: exp_result1}
\end{table*}

\section{Experiments}
\label{sec: exp_recognition}
This section shows the experiments on the observation set that includes the expert-proofread recognition results. Specifically, we conduct detailed analyses of recognition errors across various cutting-edge MLLMs and examine their relationship to downstream auto-grading performance.

\subsection{Experiment Setup}
In this experiment, we analyze the recognition capabilities of different cutting-edge MLLMs on students’ handwritten solutions in the observation set of \dsname. Specifically, we use the LLM-as-a-judge detector (introduced and validated in Section \ref{sec: Recognizing the Handwritten Content}) to identify recognition errors in the ``MLLM recognized text'' as illustrated in Figure \ref{fig:exp_overall_pipeline}. For each model, we count the number of recognition errors, categorize their types, and examine their relationship with downstream grading outcomes. All grading results are obtained using the vanilla grading pipeline in Figure \ref{fig:exp_overall_pipeline}.

We evaluate five closed-source commercial models (\texttt{Gemini-3-Pro-Preview}, \texttt{Gemini-2.5-Pro}, \texttt{Qwen3-VL-PLUS}, \texttt{Claude-4.5-Sonnet}, and \texttt{GPT-5.1}) and one open-source model (\texttt{Qwen3-VL-} \texttt{8B-Thinking}) as MLLM recognizers in Figure~\ref{fig:exp_overall_pipeline}. Additionally, we include an oracle baseline in which expert-proofread transcriptions are provided to the LLM grader, representing grading performance under perfect recognition. We use \texttt{GPT-5.1} as the LLM grader in all settings. More details can be found in Appendix \ref{experiment_details}.

\noindent\textbf{Comparison With Human Grader} In addition, we include the grades assigned by the graduate teaching assistant (denoted as \texttt{Graduate} in Table \ref{tab: exp_result1}) in the data collection course. This serves as a strong baseline, enabling us to assess how far current cutting-edge MLLM-based auto-graders have progressed in handling relatively challenging university-level problems.

\begin{figure*}[t]
    \centering
    \includegraphics[width=0.95\linewidth]{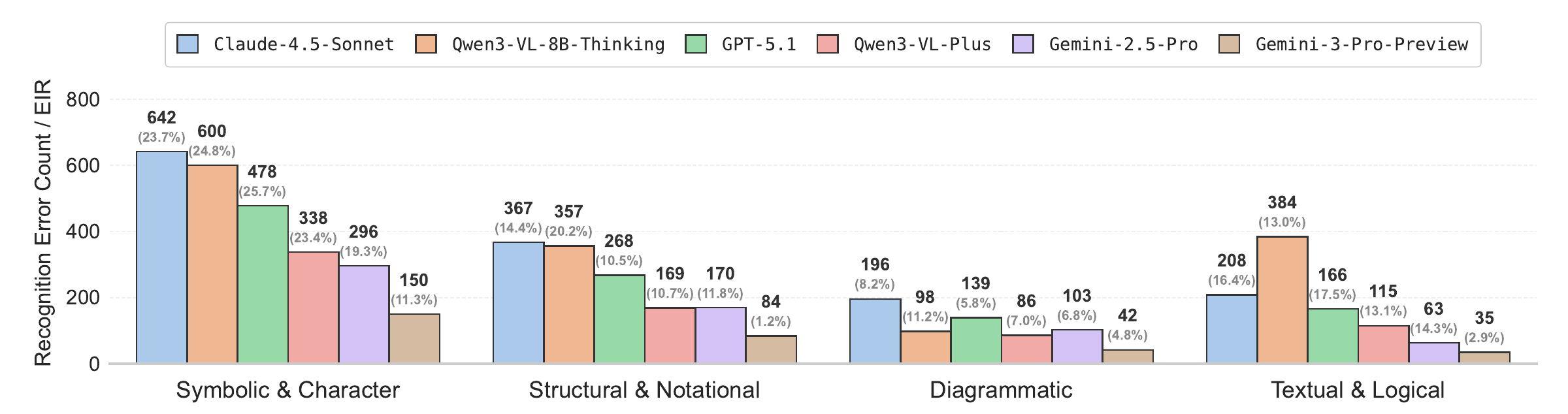} 
    \caption{Comparisons of different MLLMs' recognition error counts over 4 error categories (defined in Table \ref{tab:error_taxonomy}) and the corresponding error impact rate (EIR) to the downstream LLM grading performance in Table \ref{tab: exp_result1}. The annotation above each bar presents the error count (line 1) and the EIR (line 2 in gray) of this model in the category.}
    \label{fig:error distribution}
\end{figure*}

\noindent\textbf{Recognition Quality and Grading Metrics} We evaluate recognition performance at two granularities: Sample Error Rate (SER) and Average Error Count (AEC). SER measures the proportion of recognized texts containing at least one recognition error, while AEC represents the item-level average number of errors per handwritten solution.

To evaluate grading performance, based on the rubric introduced in Table~\ref{tab:rubric}, we measure the agreement between the LLM and expert grading reports at three levels:
(1) Binary Agreement: Evaluates whether the model correctly identifies the presence of any student mistakes;
(2) Type Agreement: Requires consistency in the flagged error types (E, M, U, C, NC);
(3) Point Agreement: The strictest metric that requires exact matches in both error types and their corresponding point deductions. For example, given an LLM grading report \texttt{\{\text{E:}-0.1pts,\text{M:}-0.3pts\}} and an expert grading report \texttt{\{\text{E:}-0.1pts,\text{M:}-0.2pts\}}, binary agreement is satisfied since both identify errors, and type agreement is also satisfied since both flag the same error types (``E'' and ``M''). However, point agreement is not satisfied due to the discrepancy in the deducted score for perspective ``M''.

\noindent\textbf{Error Impact Analysis} To further investigate the propagation of errors from visual recognition to the final grading, we introduce the Error Impact Rate (EIR). This metric quantifies the proportion of item-level recognition errors that directly lead to a downstream grading discrepancy. It is calculated as the number of recognition errors that result in a grading error divided by the total number of item-level recognition errors identified.

\subsection{Results and Findings}
The experimental results in Table \ref{tab: exp_result1} reveal several key insights. First, the findings empirically support our argument that task-centric evaluation alone fails to expose the full spectrum of visual recognition errors. Although automated grading is a multi-faceted downstream task, it inherently masks a significant portion of recognition failures (as shown in Figure \ref{fig:intro_data_showcase}). For instance, while \texttt{Gemini-3-Preview} exhibits a Sample Error Rate (SER) of 37.62\%, its Error Impact Rate (EIR) is merely 7.60\%. Nevertheless, the recognition errors have amplified negative impacts on the downstream fine-grained auto-grading tasks\textcolor{red}. According to Table \ref{tab: exp_result1}, with SER and EIR increasing from \texttt{Gemini-3-Preview} to \texttt{Qwen3-VL-8B-Thinking}, the grading performance gaps increase from 12.86\% in the binary agreement to 17.16\% and 17.35\% in the type and point agreements which require the correct identification of error types and precise point deductions. Predictably, these recognition errors may become increasingly problematic as AI-enabled educational systems \cite{chen2025wip} evolve to provide more detailed, step-by-step feedback, thereby underscoring the need of our upstream diagnostic approach. Additionally, as the overall recognition deteriorates, the EIR increases from 7.60\% to 19.60\%. These trends imply that stronger visual understanding captures more contextually significant information from student work, thereby reducing the likelihood of catastrophic error propagation.

Second, when comparing LLM-based graders with the human graduate grader, we observe a clear divergence between coarse-grained and fine-grained assessments. In terms of binary agreement, state-of-the-art MLLMs (e.g., the \texttt{Gemini} series) outperform the graduate grader, who may exhibit human rater leniency\footnote{The phenomenon whereby human graders may overlook construct-irrelevant variances (e.g., minor notation irregularities) if the student's core logic and final answer are correct.} \cite{pedagogical_filtering}. However, the human graduate grader demonstrates more consistent performance across all levels of rubric granularity, causing the advantage of MLLMs to diminish under stricter criteria. Specifically, for type- and point-level agreement, even the strongest MLLMs lag behind the human grader, with non-negligible performance gaps. This indicates that current MLLMs still struggle to reliably capture and reason over critical, fine-grained details in handwritten solutions, highlighting a persistent challenge in deploying trustworthy LLM-based systems for precision-critical educational evaluation.

\noindent\textbf{Recognition Error Taxonomy}
Based on the taxonomy defined in Table \ref{tab:error_taxonomy}, we categorize the identified recognition errors to analyze their distribution and impact, as summarized in Figure \ref{fig:error distribution}. Unsurprisingly, recognition errors are most prevalent at the surface level, particularly in the \emph{Symbolic \& Character} category, while more complex \emph{Diagrammatic} and \emph{Textual \& Logical} errors are less frequent. Beyond this, we observe that different MLLMs exhibit heterogeneous visual recognition capabilities. For instance, despite its smaller scale, \texttt{Qwen3-VL-8B-Thinking} demonstrates competitive performance in the \emph{Diagrammatic} category (98 errors), even surpassing larger models like \texttt{Gemini-2.5-Pro} (103 errors). Conversely, commercial models generally excel in the \emph{Textual \& Logical} category, showcasing their superior ability to interpret high-level semantic relations and textual explanations.

In addition, the Error Impact Rates (EIRs) in Figure \ref{fig:error distribution} reveal the sensitivity of grading to different error types. Notably, surface-level errors in \emph{Symbolic \& Character} have a more direct impact on grading outcomes, with EIRs consistently around 20\%. In contrast, identification failures in more complex \emph{Diagrammatic} and \emph{Textual \& Logical} categories generally exhibit lower EIRs, frequently falling below 10\%. This suggests that current auto-grading pipelines remain heavily reliant on symbolic matching rather than structural or diagrammatic reasoning. However, while diagrammatic errors appear less ``impactful'' for simple grading, they remain catastrophic for downstream STEM-specific tasks such as circuit-to-netlist generation \citep{rachala2022hand} or complex problem solving \citep{chen2025enhancing, skelic2025circuit}. These findings underscore that achieving truly robust and comprehensive STEM handwriting understanding still remains a significant challenge.


\section{Case Study}
\label{sec: case_study}
To assess the practical utility of our recognition error analysis, we conduct a deployment-oriented case study that integrates our empirical recognition error diagnostics into a realistic human-in-the-loop grading workflow (i.e., the ``regrading module'' in Figure~\ref{fig:exp_overall_pipeline}). Specifically, we first collect identified recognition errors from the observation set and summarize their patterns. These patterns are then used to prompt the LLM to identify potential errors within the MLLM-recognized student solutions in the test set, which simulate unseen scenarios in practice. Meanwhile, considering that the distinction between genuine recognition errors and students' own mistakes can sometimes be ambiguous (due to handwriting or logical issues), we further ask the LLM to provide a confidence level (``high’’ or ``low’’) for each identified error. Solutions flagged with low-confidence errors are routed directly to the course teaching assistant for manual grading, while the remaining solutions are re-graded by the LLM grader given the detector’s error report. More details can be found in Appendix \ref{sec: case study detail}.

\begin{table}[t]
\centering
\resizebox{\columnwidth}{!}{%
    \renewcommand{\arraystretch}{1.1}
    \setlength{\tabcolsep}{1pt} 
    
    \begin{tabular}{w{l}{2.25cm} l ccc w{c}{1.2cm} w{c}{1.2cm}} 
    \toprule
    \multirow{2}{*}{\textbf{Workflow}} & \multirow{2}{*}{\textbf{Visual Recognizer}} & \multicolumn{3}{c}{\textbf{Grading Perf. (\%) $\uparrow$}} & \multicolumn{2}{c}{\textbf{Regrading Rate (\%)}} \\
    \cmidrule(lr){3-5} \cmidrule(lr){6-7}
     & & \textbf{Binary} & \textbf{Type} & \textbf{Point} & \textbf{LLM} & \textbf{Human} \\
    \midrule
    \multirow{2}{*}{Vanilla} 
     & \texttt{Gemini-2.5-Pro} & \cellcolor{softred} 85.02 & \cellcolor{softred} 74.91 & \cellcolor{softred} 69.91 & -- & -- \\
     & \texttt{GPT-5.1}        & \cellcolor{softred} 82.34 & \cellcolor{softred} 72.23 & \cellcolor{softred} 66.87 & -- & -- \\
     \midrule
    \multirow{2}{*}{w/ Regrading} 
     & \texttt{Gemini-2.5-Pro} & \cellcolor{softgreen} 86.48 & \cellcolor{softgreen} 77.34 & \cellcolor{softgreen} 74.42 & 20.6 & 3.3 \\
     & \texttt{GPT-5.1}        & \cellcolor{softgreen} 86.60 & \cellcolor{softgreen} 78.93 & \cellcolor{softgreen} 75.76 & 25.1 & 4.4 \\
    \bottomrule
    \end{tabular}
}
\vspace{-0.5em}
\caption{Performance and regrading statistics \colorbox{softred}{before} and \colorbox{softgreen}{after} applying the regrading module for student solutions in the test set. \textbf{LLM} and \textbf{Human} columns denote the percentage of solutions assigned to LLM and teaching assistant for regrading, respectively.}
\vspace{-0.5em}
\label{tab:selected_grading_comparison}
\end{table}

\noindent\textbf{Deployment Setup}
We use the ``vanilla grading pipeline'' in Figure \ref{fig:exp_overall_pipeline} for the first round student solution grading, where \texttt{GPT-5.1} and \texttt{Gemini-2.5-Pro} are tested as the visual recognizer separately. Subsequently, we apply the ``regrading module'' to regrade all solutions that received point deductions in the first round. This design is motivated by both error characteristics and deployment costs: recognition errors primarily result in false positive grading penalties, and cases with point deductions generally constitute a small fraction of student solutions. The LLM grader, LLM re-grader, and recognition error detector are all implemented using \texttt{GPT-5.1}.

\noindent\textbf{Performance Gain from Regrading} Table~\ref{tab:selected_grading_comparison} indicates that the ``regrading module'' significantly improves grading performance, particularly in type and point agreement, which necessitate precise alignment of fine-grained details. Notably, with fewer than 5\% solutions requiring human intervention, the regrading achieves performance comparable to the setting where human experts serve as visual recognizers for all solutions in Table~\ref{tab: exp_result1}. This demonstrates that despite the low error impact ratio, targeted recognition error suppression remains indispensable for achieving expert-level accuracy in complex STEM evaluation.

\section{Conclusions}
In this work, we introduce \dsname, a real-world benchmark comprising over 1,300 authentic university-level STEM handwritten student solutions. Leveraging expert-verified verbatim handwritten solution transcriptions and grading reports, we comprehensively evaluate the visual understanding capability of MLLMs and their cascading impact on downstream auto-grading. The evaluation uncovers a substantial volume of latent recognition errors that may be exposed when rubrics become stricter and more professional, highlighting the significant challenges in deploying reliable MLLM-based educational applications. Furthermore, in response to these challenges, we propose and validate a human-in-the-loop grading workflow to mitigate the impact of recognition failures, which significantly enhances system robustness with minimal human intervention. We hope our work can serve as a cornerstone for developing more trustworthy AI-enabled educational technologies.

\section{Data Scope: Beyond Circuit Analysis}
Note that although the student handwritten data in \dsname was collected from a university-level course on circuit analysis, this does not imply that our dataset merely serves as a proxy for a narrow subset of STEM challenges. In fact, solving the problems in our dataset requires interleaved reasoning across multiple foundational disciplines, including calculus, differential equations, and physics, which are formal prerequisites for the course \cite{sun2025data}. In Figures \ref{fig: example_1_for_data_scope_clarificiation} and \ref{fig: example_2_for_data_scope_clarificiation}, we present two recognized handwritten solutions corresponding to Problems 9.8-2 and 10.6-4 from the textbook by \citet{svoboda2013introduction}. These examples illustrate that the \dsname dataset is also relevant to foundational STEM areas such as physics, differential equations, calculus, and linear algebra. Due to copyright restrictions, we do not include the original problem statements; instead, we provide brief summaries as follows:

i) The first example is a recognized handwritten solution to Problem 9.8-2 in \citet{svoboda2013introduction}. The circuit diagram includes two resistors, one capacitor, one inductor, and an independent voltage source with voltage $2u(t) - 1$. The task is to determine the current $i(t)$ through the inductor for $t > 0$. The student’s solution involves fundamental circuit laws, including Kirchhoff’s Current Law (KCL) and Kirchhoff’s Voltage Law (KVL), as well as the formulation and solution of a \textbf{second-order linear differential equation}, the \textit{derivation of the characteristic equation in the Laplace domain}, and the representation of the final current as the sum of two exponential terms and a constant term. These elements indicate that the benchmarking results derived from our dataset are not limited to circuit analysis alone.

ii) The second example is a recognized handwritten solution to Problem 10.6-4 in \citet{svoboda2013introduction}. The circuit diagram contains an independent voltage source represented by the phasor $48\angle 75^\circ$ and five impedance elements with values $40 + j15 \Omega$, $25 - j50 \Omega$, $32 + j16 \Omega$, $-j50 \Omega$, and $j40 \Omega$. The task is to determine the values of three phasor mesh currents. The student’s solution involves formulating three mesh-current equations that constitute a \textit{system of linear equations}. When expressed in matrix form, this system includes complex-valued coefficients. Recognition and evaluation of this solution therefore require not only knowledge of circuit analysis but also proficiency in linear algebra.

We hope this clarification of our dataset's scope helps future researchers better understand and utilize the \dsname dataset.




\section{Limitations}
This work focuses on the analysis on handwritten solution visual understanding capability in a university-level circuit analysis course. Although such a course represents a comprehensive STEM subject that integrates concepts from mathematics, physics, and engineering, the problem set and accompanying diagrams are mainly centered on circuit-related analysis. Other diagram types strongly associated with abstract mathematics (e.g., complex geometric graphs) are underrepresented in our dataset. Therefore, the conclusions drawn in this study may not fully generalize to handwritten recognition tasks involving substantially different diagram modalities.

In addition, in our analysis of how upstream recognition errors propagate to downstream tasks, we mainly consider auto-grading as the primary downstream application. Although auto-grading is a practical and representative task in educational settings, we do not investigate the impact of recognition errors on other downstream tasks, such as visual question answering (VQA), which may exhibit different sensitivity patterns to recognition inaccuracies.

Finally, the grading labels and rubrics in our dataset are provided by few doctoral experts with extensive familiarity with the course content. However, human evaluation and rubric-crafting in open-ended STEM problem-solving are inherently subjective to some extent. Despite our efforts to ensure consistency through detailed grading criteria and rigorous verification grounded in mathematical logic and domain expertise, potential subtle subjective biases may still exist in the grading scores.

Overall, we view these limitations as opportunities for future work to extend the analysis to broader subject domains, additional downstream tasks, and alternative assessment protocols.

\section{Ethics Considerations}
\textbf{Data Collection} Prior to data collection, all participating students signed informed consent forms, authorizing the collection and usage of their homework solutions and grading records for research and publication purposes. All data were anonymized to remove Personally Identifiable Information (PII) of students (e.g., we identify students in our dataset via ``student + index'' (student 1, student 6, etc) instead of their true names for privacy protection.), and the collection was approved by the Institutional Review Board (IRB).

\noindent\textbf{Copyright and Licensing} The 62 questions used in this study are sourced from the textbook \cite{svoboda2013introduction}, with their corresponding IDs listed in Table \ref{tab: question_involved}. To comply with copyright regulations, \dsname does not distribute and publicize the original problem statements, figures, or image-based hints. Researchers who wish to replicate our experiments involving the original question content are advised to consult the textbook themselves using the provided IDs. However, for all questions listed in Table \ref{tab: question_involved}, we have independently developed the reference solutions and comprehensive grading rubrics (as exemplified in Figure \ref{prompt: rubric example}). These annotations are original contributions and are fully included in \dsname.

\noindent\textbf{Reliability Risks in Educational Settings} As analyzed in our work, MLLMs adopted in current auto-grading systems present unreliable recognition qualities on university-level STEM student solutions, which could propagate to downstream grading performance and thereby lead to unfair or incorrect assessments if deployed without sufficient human oversight. We emphasize that our dataset and analysis are intended for diagnostic and research purposes, and that any practical deployment should include human-in-the-loop verification and error detection mechanisms, as shown in the case study.

\section{Acknowledgments}
The authors appreciate the support provided by the School of Electrical and Computer Engineering and the College of Engineering at the institution where the study was conducted.

\bibliography{custom}

\clearpage

\appendix

\section{Related Work}
\subsection{Handwriting Text Recognition}
Despite the prevalence of MLLMs, they still face several challenges. One notable issue is the difficulty in accurately recognizing students' handwriting and converting it into clean, machine-readable text. Although Optical Character Recognition (OCR) \cite{hamad2016detailed} and visual understanding techniques can reliably process standard text and simple, one-line mathematical expressions \cite{Wu_Du_Li_Zhang_Yang_Ren_Hu_2022}, their performance degrades when confronted with complex deductions, hand-drawn auxiliary plots, and uncommon symbol combinations. This leads to low confidence in the automated evaluation of student work \cite{yang2025pensieve, caraeni2024evaluating}, thereby increasing the need for manual verification \cite{yang2025pensieve}. Consequently, it is meaningful to analyze in detail the impact of recognition errors on auto-grading systems and to develop targeted solutions for these shortcomings.

\subsection{Handwriting Text Recognition Correction}
Handwriting Text Recognition (HTR) correction plays a crucial role in identifying and rectifying recognition errors, thereby improving the overall performance of the recognition pipeline. For example, \citet{pavlopoulos2023detecting} demonstrated that a sequential ``detect-then-rectify'' approach could enhance Greek text recognition accuracy to 97\%. Similarly, \citet{chen2023language} employed an external LLM to automatically correct mathematical formulas by leveraging LaTeX grammar and semantic context, achieving an approximate 6\% improvement in expression recognition accuracy on the CROHME dataset \cite{2014crohme}.

Methods for error detection vary, including BERT-based models \cite{peng2021mathbert} and confidence-based approaches \cite{gupta2021unsupervised}. However, their effectiveness remains underexplored in the context of complex student handwriting, especially when accounting for practical constraints such as minimizing API call frequency and ensuring high-recall error detection (i.e., identifying as many errors as possible). This paper introduces a specialized error detection and correction pipeline tailored to student handwriting, developed and evaluated within the context of a university-level circuit analysis course, as detailed in the case study in Section \ref{sec: case_study}.

\subsection{Automated Grading}
Recent auto-grading agents primarily leverage the capabilities of foundation models such as Gemini and GPT. For handwriting input, some researchers have explored end-to-end grading approaches that process handwritten submissions directly \cite{caraeni2024evaluating, khrulev2025check, nath2025can}. However, the opaque nature and relatively lower performance of these models render them unreliable at the current stage. Alternatively, a more interpretable approach involves first recognizing the student’s handwritten solution and converting it into a structured text format (e.g., LaTeX or Markdown), which is then used as input for the grading model along with a prompt and predefined rubric. For instance, \citet{liu2024ai} utilized GPT-4 to automatically grade a university-level mock mathematics exam. \citet{kortemeyer2024grading} combined GPT-4V and Mathpix to evaluate student performance on a high-stakes thermodynamics exam. Similarly, \citet{parsaeifard2025automated} developed a pipeline to automatically grade economics-related mathematics courses at a Swiss distance university. \citet{yang2025pensieve} introduced an online auto-grading platform\footnote{See \url{https://www.pensieve.co/schools}.} that supports various academic institutions across subjects such as Computer Science, Mathematics, and Chemistry.

\begin{figure*}[t] 
    \begin{promptbox}{Prompt for categorizing the recognition errors based on the error taxonomy}
As an expert in Electronic Engineering education and OCR quality assessment, categorize the following recognition error into one of the four major types:

Category 1: Symbol & Character Errors (Low-level OCR)
- Includes: Misreading numbers/letters (e.g., '20000' as '2a000'), missing negative signs, operator misrecognition, and unit/prefix errors (e.g., '\muF' as 'mF' or 'MF').

Category 2: Structural & Mathematical Notation Errors
- Includes: Broken formula structures (e.g., fractions split across lines), missing parentheses, inconsistent variables (e.g., 'vc' becoming 'vo' mid-derivation), and dimensional inconsistencies.

Category 3: Visual & Diagrammatic Reasoning Errors
- Includes: Incorrect circuit topology (e.g., identifying parallel as series), misaligning attributes to wrong components, reversed polarity/direction, and hallucinations regarding diagram elements.

Category 4: Textual & Logical Flow Errors
- Includes: Omission of critical steps, mislabeling physics laws (e.g., writing KVL when KCL was used), and boundary condition mismatches (e.g., 't < 0' misread as 't > 0').

---
Error Item Content:
{item_content}
---

Task: Output ONLY the category number (1, 2, 3, or 4). Do not include any other text.
    \end{promptbox}
\caption{Prompt used to categorize recognition errors according to the taxonomy defined in Table \ref{tab:error_taxonomy}, where the four categories correspond to ``Symbol \& Character'', ``Structural \& Notational'', ``Diagrammatic'', and ``Textual \& Logical'' errors.}
\label{prompt: error_taxonomy}
\end{figure*}

\section{Dataset Details and Ethics Statements}
\label{sec: dataset detail}
The \dsname dataset consists of the following components: (1) \dscount authentic student handwritten solutions collected from a university-level STEM course during the Spring 2025 term; (2) 62 reference answers and corresponding grading rubrics; (3) 513 expert-crafted, near-verbatim transcriptions of student solutions within the observation set; and (4) detailed records of recognition errors, categorized according to the taxonomy presented in Table \ref{tab:error_taxonomy}. Further descriptions of these components are provided below.

\noindent (1) \textbf{Student Handwritten Solution Collection.} As summarized in Table \ref{tab:dataset_summary}, data were collected from 29 students enrolled in the course. Prior to data collection, all participants signed an informed consent form permitting the use of their homework submissions and grading scores for research purposes. All data were anonymized to remove Personally Identifiable Information (PII), and the collection process was approved by the Institutional Review Board (IRB). Handwritten assignments and corresponding grading reports were gathered throughout the Spring 2025 term. Notably, some students occasionally submitted assignments in LaTeX or other digital formats; these were excluded from the dataset. As a result, not all students in \dsname have handwritten solutions for every question.

\noindent (2) \textbf{Questions, Reference Solutions, and Rubrics.} The IDs of all 62 questions, sourced from the textbook \cite{svoboda2013introduction}, are listed in Table \ref{tab: question_involved}. Due to copyright restrictions, \dsname does not include the original problem statements, figures, or image-based hints. Researchers wishing to replicate experiments involving the original content are encouraged to consult the textbook using the provided question IDs. For all questions listed in Table \ref{tab: question_involved}, we developed corresponding reference answers and comprehensive grading rubrics, both of which are included in \dsname. An example rubric is shown in Figure \ref{prompt: rubric example}.

\noindent (3) \textbf{Expert-Rectified Student Solution Recognitions.} This component is described in greater detail in Appendix \ref{sec: crafting_gold_standard_recog_text} and was previously introduced in Section \ref{sec: Recognizing the Handwritten Content}.

\noindent (4) \textbf{Recognition Errors and Taxonomy.} As detailed in Section \ref{sec: Recognizing the Handwritten Content} and Appendix \ref{sec: crafting_gold_standard_recog_text}, human experts identified recognition errors in the transcriptions produced by \texttt{Gemini-2.5-Pro} on the observation set within \dsname. These errors were then categorized using the taxonomy defined in Table \ref{tab:error_taxonomy}, with classification performed by prompting \texttt{Gemini-2.5-Pro} using the prompt shown in Figure \ref{prompt: error_taxonomy}. In addition, we will release the recognition errors identified across all evaluated models (as shown in Figure \ref{fig:error distribution}), flagged by the LLM-based error detector and organized according to the same taxonomy, to support further research.

\begin{table*}[t]
    \centering
    \small
    \renewcommand{\arraystretch}{1.2}
    
    \begin{tabularx}{0.9\linewidth}{c|X}
        \toprule
        \textbf{Chapter Index} & \multicolumn{1}{c}{\textbf{Problem Indexes}} \\ 
        \midrule
        1  & \texttt{P1.5-2}, \texttt{P1.5-3} \\
        2  & \texttt{P2.4-6}, \texttt{P2.4-7}, \texttt{P2.5-1}, \texttt{P2.5-2} \\
        3  & \texttt{P3.2-1}, \texttt{P3.2-2}, \texttt{P3.2-6}, \texttt{P3.4-4}, \texttt{P3.6-1} \\
        4  & \texttt{P4.2-2}, \texttt{P4.2-5}, \texttt{P4.3-3}, \texttt{P4.4-1}, \texttt{P4.4-6}, \texttt{P4.5-2}, \texttt{P4.6-2}, \texttt{P4.7-1}, \texttt{P4.7-2}, \texttt{P4.7-6} \\
        6  & \texttt{P6.3-2}, \texttt{P6.3-3}, \texttt{P6.4-1}, \texttt{P6.4-2}, \texttt{P6.4-4}, \texttt{P6.4-5}, \texttt{P6.4-6} \\
        7  & \texttt{P7.2-2}, \texttt{P7.2-4}, \texttt{P7.3-2}, \texttt{P7.4-1}, \texttt{P7.4-3}, \texttt{P7.5-4}, \texttt{P7.5-5}, \texttt{P7.6-2}, \texttt{P7.7-2}, \texttt{P7.8-2} \\
        8  & \texttt{P8.3-1}, \texttt{P8.3-2}, \texttt{P8.3-4}, \texttt{P8.3-5}, \texttt{P8.3-9}, \texttt{P8.3-10}, \texttt{P8.4-1}, \texttt{P8.7-3}, \texttt{P8.7-5} \\
        9  & \texttt{P9.2-1}, \texttt{P9.3-3}, \texttt{P9.4-1}, \texttt{P9.4-2}, \texttt{P9.5-1}, \texttt{P9.6-1}, \texttt{P9.7-2} \\
        10 & \texttt{P10.3-2}, \texttt{P10.3-3}, \texttt{P10.4-1}, \texttt{P10.4-6}, \texttt{P10.6-1}, \texttt{P10.6-4}, \texttt{P10.7-1}, \texttt{P10.10-4} \\ 
        \bottomrule
    \end{tabularx}
    \caption{Index of problems from \cite{svoboda2013introduction} corresponding to the data included in \dsname.}
    \label{tab: question_involved}
\end{table*}

\section{Expert Handwriting Recognition}
\label{sec: crafting_gold_standard_recog_text}
This section outlines the procedure for generating expert-verified handwritten recognition results, referred to as the ``gold standard'', for student solutions. For all data in the \dsname observation set, we initially employed \texttt{Gemini-2.5-Pro} to recognize and transcribe the handwritten submissions into Markdown format, using the prompt shown in Figure \ref{prompt: transcription_common}. Subsequently, human experts carefully reviewed each transcription against the original handwritten images. Any recognition errors identified during this verification process were manually corrected by the experts. The resulting set of transcriptions serves as the gold standard and is assumed to be error-free for the experiments reported in Table \ref{tab: exp_result1}.

\begin{figure}[h]
\includegraphics[width=1\linewidth]{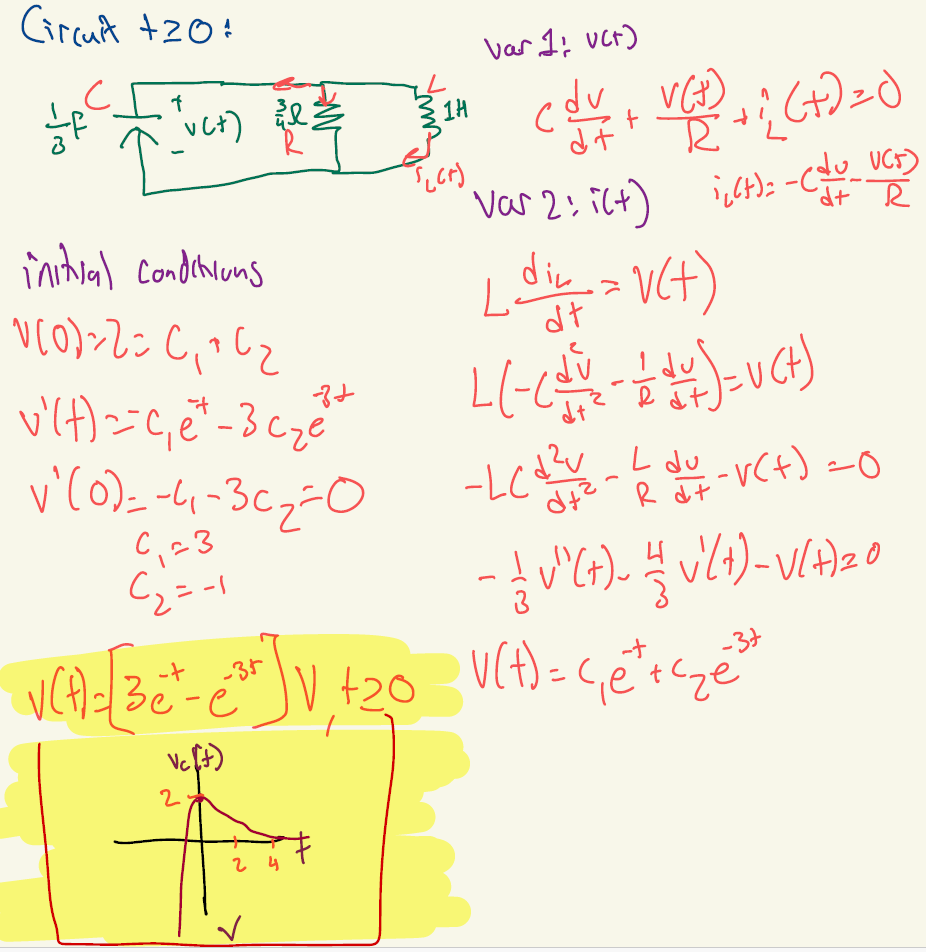}
\caption{Screenshot of a handwritten solution from Student 16 for question P9.4-2, included in the observation set.}
\label{fig: student example P9.4-2}
\end{figure}

To better illustrate the expert rectification process, we provide a representative example. Figure~\ref{fig: student example P9.4-2} presents the original student handwritten submission; Figure \ref{prompt: student example P9.4-2 gemini recog} shows the initial transcription generated by \texttt{Gemini-2.5-Pro}; and Figure \ref{prompt: student example P9.4-2 expert recog} displays the expert-rectified version based on the model’s output. In these figures, recognition errors in the original transcription are highlighted in red, while the corresponding corrections in the expert-verified version are marked in green.

\begin{figure}[h]
\includegraphics[width=1\linewidth]{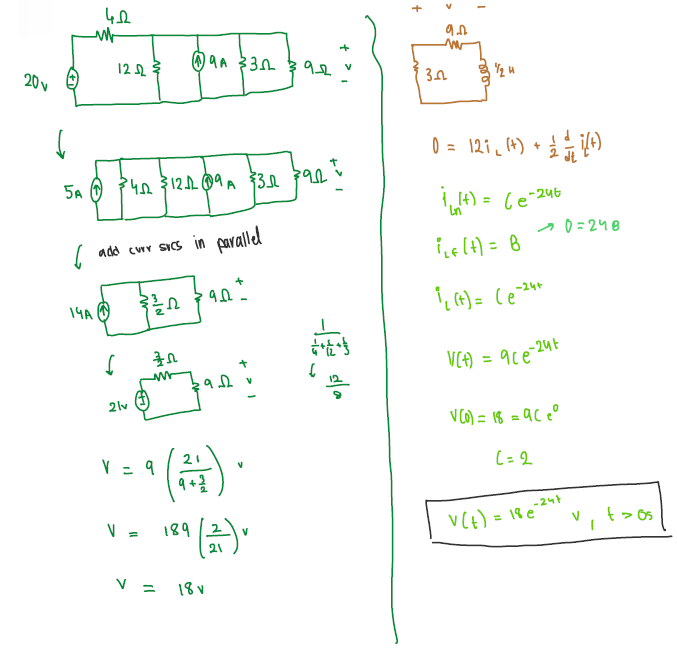}
\caption{Screenshot of a student handwritten solution for question P8.3-10 from the test set. This example is used as the one-shot demonstration in the LLM-based recognition error detector described in Section \ref{sec: Recognizing the Handwritten Content} for evaluating the observation set.}
\label{fig: student example P8.3-10}
\end{figure}

\section{Experiment Details}
\label{experiment_details}
\subsection{Automated LLM-Enabled Recognition Error Detector}

This section describes the implementation of the LLM-enabled automated recognition error detector, as introduced in Section \ref{sec: Recognizing the Handwritten Content}. Leveraging the expert-verified transcriptions as the ``gold standard'', we use \texttt{Gemini-2.5-Pro} to detect discrepancies between the recognition outputs produced by various MLLMs and the corresponding reference texts. The prompt used for this comparison process is illustrated in Figure~\ref{prompt: recognition effect comparison}, and consists of the following components:

\begin{itemize}
\item \textbf{\texttt{target\_content}}: The Markdown transcription generated by the target MLLM for a specific student handwritten solution.
\item \textbf{\texttt{label\_content}}: The expert-verified Markdown transcription of the same student solution, generated through the pipeline described in Section \ref{sec: crafting_gold_standard_recog_text}. This serves as the gold-standard reference for evaluating recognition accuracy.
\item \textbf{\texttt{examples[`target']}}: A sample transcription generated by \texttt{GPT-5.1} for the student submission shown in Figure \ref{fig: student example P8.3-10}, with the transcribed output illustrated in Figure \ref{prompt: student example P8.3-10 GPT recog}. To ensure fair evaluation (e.g., in Tables \ref{tab:human_vs_automatic} and \ref{tab: exp_result1}) and prevent data leakage, this example, corresponding to question P8.3-10 from Student 19, was selected from the test set, which lies outside the \dsname observation set.
\item \textbf{\texttt{examples[`gt']}}: The expert-rectified transcription corresponding to the student submission shown in Figure~\ref{fig: student example P8.3-10}, as illustrated in Figure~\ref{prompt: student example P8.3-10 expert recog}.
\item \textbf{\texttt{examples[`result']}}: A list of sample recognition errors and their corresponding corrections, derived by comparing \texttt{examples[`target']} with \texttt{examples[`gt']}. This list is shown in Figure \ref{prompt: example_about_recognition_error}, and the detector's output follows the same structure.
\item \textbf{\texttt{examples[`evidence']}}: Detailed justifications for each recognition error identified in \texttt{examples[`target']}, based on comparisons with \texttt{examples[`gt']}. These justifications are illustrated in Figure \ref{prompt: explanation_for_example_about_recognition_error}.
\end{itemize}

\noindent This prompt utilizes a one-shot strategy based on a representative example from the test set. Specifically, we intentionally paired a transcription produced by \texttt{GPT-5.1} with an expert-rectified ground truth originally derived from \texttt{Gemini-2.5-Pro}. By incorporating outputs from different MLLMs in the one-shot example, the detector is exposed to the natural variability in how models represent the same visual content, for instance, through differing LaTeX notations for equivalent formulas or alternative phrasings for diagrammatic descriptions. This cross-model pairing enhances the robustness of the LLM-based detector by helping it distinguish between benign structural or stylistic differences and true recognition errors.

\subsection{Vanilla Grading Pipeline}
The ``vanilla grading pipeline'', illustrated in Figure \ref{fig:exp_overall_pipeline}, comprises two main stages: (1) transcription of the student handwritten solution by an MLLM, such as \texttt{Gemini-2.5-Pro}, and (2) automated evaluation by an LLM-based grader. In the transcription stage, we utilize different MLLMs to convert handwritten student responses into structured text, guided by prompts specific to each model series (see Figures \ref{prompt: transcription_common} and \ref{prompt: transcription_qwen}).

In the second stage, the LLM grader (i.e., \texttt{GPT-5.1}), configured with a ``high'' reasoning effort setting, evaluates the transcribed content through a two-round dialogue process. In the first round, the grader employs the prompt shown in Figure \ref{prompt: 1st round grading} to assess the submission based on four question-specific attributes: \texttt{problem\_statement}, \texttt{images\_hint}, \texttt{final\_answer} (the reference solution), and \texttt{rubric\_snippet}. Within \dsname, we provide the latter two components, developed by domain experts, while the \texttt{problem\_statement} and \texttt{images\_hint} can be retrieved from the textbook \citep{svoboda2013introduction} using the corresponding question indexes. The \texttt{student\_markdown} field contains the transcribed student solution, which may originate from an MLLM (e.g., Figure \ref{prompt: student example P8.3-10 GPT recog}) or a human expert, depending on the experimental setting. After the first round, the grader outputs an overall score deduction report (e.g., \texttt{\{E: -0.04 pts, C: -0.01 pts}\}).

\begin{figure}[h]
\includegraphics[width=1\linewidth]{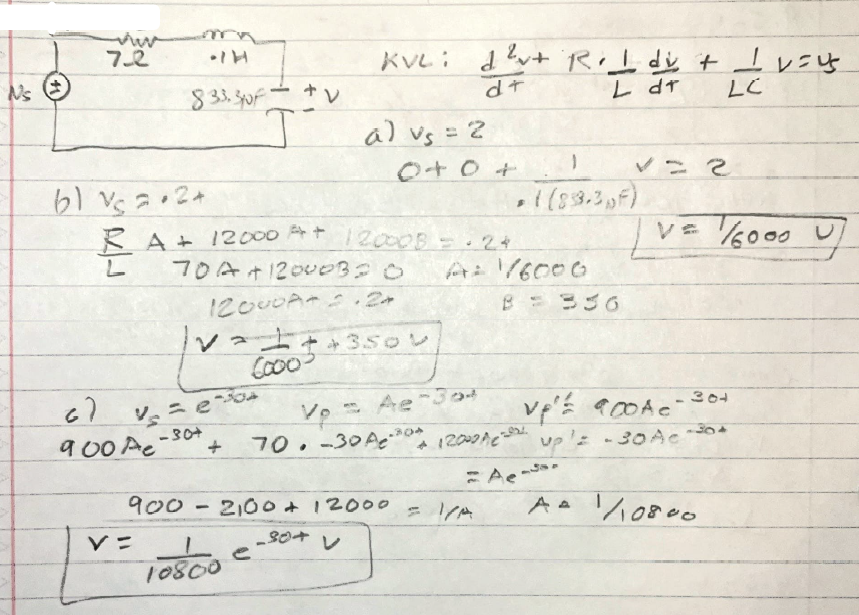}
\caption{Screenshot of a student handwritten solution for question P9.7-2. A grading report example for this submission, generated by the LLM grader, is presented in Figure \ref{prompt: grading report sample (P9.7-2)}, based on the rubric shown in Figure \ref{prompt: rubric example}.}
\label{fig: student example P9.7-2}
\end{figure}

In the second round, the LLM grader is prompted (as shown in Figure \ref{prompt: 1st round grading_system}) to provide detailed justifications for the score deductions identified in the first round. The input to this stage includes the \texttt{raw\_scoring\_json} (e.g., \texttt{\{E: -0.04 pts, C: -0.01 pts\}}) generated earlier. This round concludes with the generation of a final grading report. A representative example is shown in Figure \ref{prompt: grading report sample (P9.7-2)}, which presents the LLM grader's feedback on the student submission in Figure \ref{fig: student example P9.7-2}, evaluated using the rubric provided in Figure \ref{prompt: rubric example}.

\begin{figure}[t]
    \begin{promptbox}{A grading report example}
E: EE1 (0.04) - The governing ODE is written with v_s on the right-hand side instead of (1/LC) v_s, so the source term is incorrectly scaled, making the ODE relating v and v_s wrong. | evidence: $$\frac{d^2v}{dt} + R \cdot \frac{1}{L} \frac{dv}{dt} + \frac{1}{LC} v = v_s$$

C: CC1 (0.01) - From 70A + 12000B = 0 and 12000A = 0.2, solving gives a small negative B, not 350; this is an algebra error in solving for the ramp-case constant B. | evidence: $$B = 350$$
    \end{promptbox}
\caption{Grading report for a student submission on question P9.7-2. The report includes score deductions in the categories of ``Equation'' (E) and ``Calculation'' (C), along with detailed explanations for each deduction and the corresponding evidence (e.g., \texttt{evidence: \$\$B = 350\$\$}) from the transcribed student solution.}
\label{prompt: grading report sample (P9.7-2)}
\end{figure}

\begin{table}[h!]
    \centering
    \begin{tblr}{
        colspec = {X[0.45,c,m] X[0.55,c,m]}, 
        width = \columnwidth,
        hline{1,2,3,4,Z} = {0.8pt}, 
    }
        \textbf{Image} & \textbf{Error Details} \\
        
        \includegraphics[width=\linewidth, valign=m]{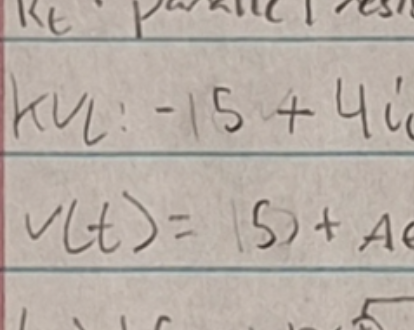} & 
        {KVL: \textcolor{green!60!black}{$-5 + 4i_1 \text{ ...}$}\\ $\downarrow$ \\KVL: \textcolor{red}{$-15 + 4i_1 \text{ ...}$}}  \\

        \includegraphics[width=\linewidth, valign=m]{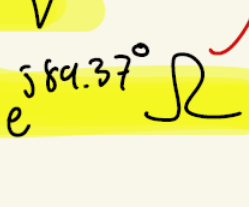} & 
        {\textcolor{green!60!black}{$e^{j89.37^{\circ}}$} \\ $\downarrow$ \\ \textcolor{red}{$e^{j84.37^{\circ}}$}} \\
    \end{tblr}
    \caption{Examples of the recognition errors in the ``Symbolic \& Character'' category in Table \ref{tab:error_taxonomy}. The incorrect recognitions are marked in \textcolor{red}{red} and the rectified versions are marked in \textcolor{green!60!black}{green}.}
    \label{tab:symbol_error_examples}
\end{table}

\subsection{Regrading Module}
\label{sec: case study detail}
This section provides additional details on the ``regrading module'' in Figure \ref{fig:exp_overall_pipeline}, which was implemented as part of the case study described in Section \ref{sec: case_study}.

\begin{table}[h!]
    \centering
    \begin{tblr}{
        colspec = {X[0.45,c,m] X[0.55,c,m]}, 
        width = \columnwidth,
        hline{1,2,3,4,Z} = {0.8pt}, 
    }
        \textbf{Image} & \textbf{Error Details} \\ 
        
        \includegraphics[width=\linewidth, valign=m]{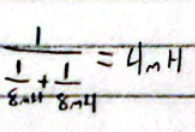} & 
        {\textcolor{green!60!black}{$\frac{1}{\frac{1}{8~\text{mH}} + \frac{1}{8~\text{mH}}} = 4~\text{mH}$}\\ $\downarrow$ \\ \textcolor{red}{$\frac{1}{8~\text{mH} + 8~\text{mH}} = 4~\text{mH}$}}  \\

        \includegraphics[width=\linewidth, valign=m]{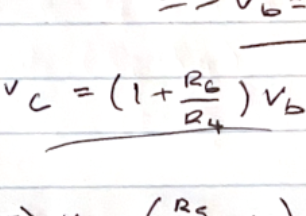} & 
        {\textcolor{green!60!black}{$v_c = (1+\frac{R_6}{R_4})v_b$} \\ $\downarrow$ \\ \textcolor{red}{$v_c = (\frac{1+R_6}{R_4})v_b$}} \\
    \end{tblr}
    \caption{Examples of the recognition errors in the ``Structural \& Notational'' category in Table \ref{tab:error_taxonomy}. The incorrect recognitions are marked in \textcolor{red}{red} and the rectified versions are marked in \textcolor{green!60!black}{green}.}
    \label{tab:eq_error_examples}
\end{table}

\begin{table}[t]
    \centering
    \begin{tblr}{
        colspec = {X[0.45,c,m] X[0.55,l,m]}, 
        width = \columnwidth,
        hline{1,2,3,4,Z} = {0.8pt}, 
    }
        \textbf{Image} & \SetCell{c} \textbf{Error Details} \\ 
        
        \includegraphics[width=\linewidth, valign=m]{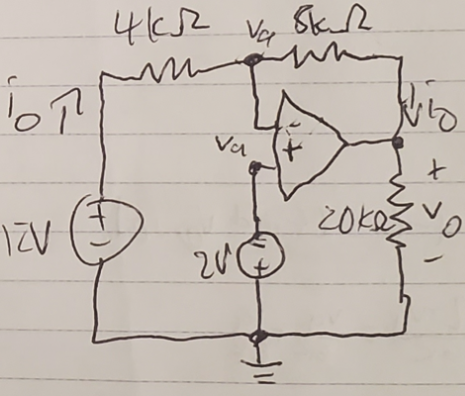} & 
        {\noindent $i_o$ follows downwards through \textcolor{green!60!black}{8~k$\Omega$ resistor}.\\ $\quad\quad\quad\quad\downarrow$ \\ $i_o$ follows downwards through \textcolor{red}{20~k$\Omega$ resistor}. }  \\

        \includegraphics[width=\linewidth, valign=m]{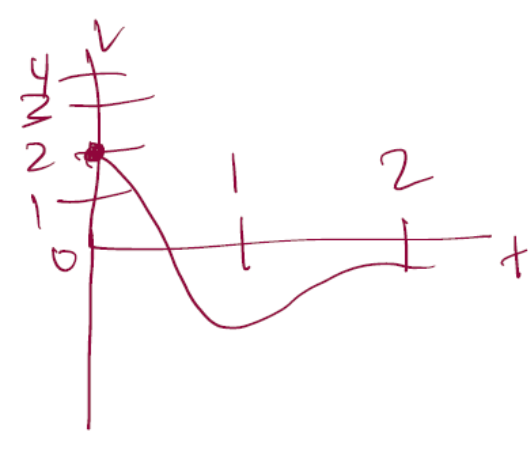} & 
        {\noindent The curve has a minimum between \textcolor{green!60!black}{$t$=0 and $t$=1}. \\ $\quad\quad\quad\quad\downarrow$ \\ The curve has a minimum between \textcolor{red}{$t$=1 and $t$=2}.} \\
        
        \includegraphics[width=\linewidth, valign=m]{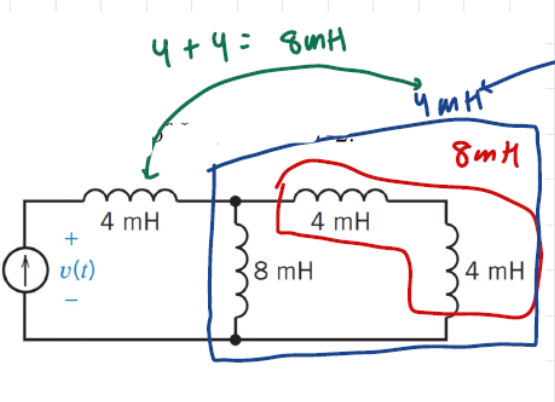} & 
        {\noindent The student circles \textcolor{green!60!black}{two 4mH inductors (top right \& rightmost) in red}. \\ 
          $\quad\quad\quad\quad\downarrow$ \\ 
          The student circles  \textcolor{red}{the rightmost 4mH inductor in red}.
        } \\
    \end{tblr}
    \caption{Examples of the recognition errors in the ``Diagrammatic'' category in Table \ref{tab:error_taxonomy}. The incorrect recognitions are marked in \textcolor{red}{red} and the rectified versions are marked in \textcolor{green!60!black}{green}.}
    \label{tab:diagrammatic_error_examples}
\end{table}

\begin{table}[t]
    \centering
    \begin{tblr}{
        colspec = {X[0.45,c,m] X[0.55,c,m]}, 
        width = \columnwidth,
        hline{1,2,3,Z} = {0.8pt}, 
    }
        \textbf{Image} & \textbf{Error Details} \\ 
        
        \includegraphics[width=\linewidth, valign=m]{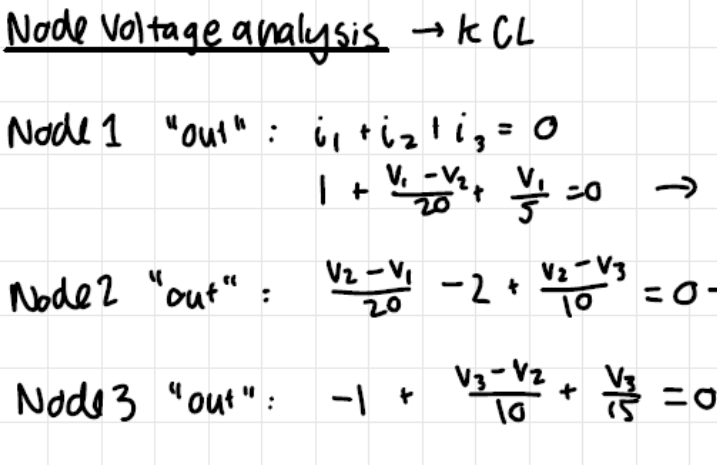} & 
        {\small Node voltage analysis $\rightarrow$ \textcolor{green!60!black}{KCL} ...\\ $\downarrow$ \\ Node voltage analysis $\rightarrow$ \textcolor{red}{KVL} ...}  \\

        \includegraphics[width=\linewidth, valign=m]{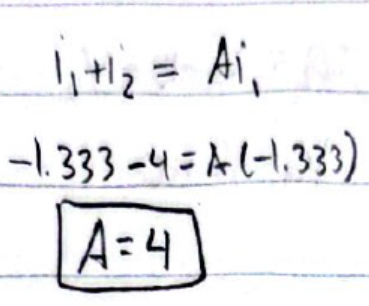} & 
        {\small 
         \textcolor{green!60!black}{$i_1+i_2=\text{A}i_1$}. \\ 
         $-1.333-4=\text{A}(-1.333)$ \\ ... \\
         $\downarrow$ \\ 
         \textcolor{red}{[Missing Equation]}. \\ 
         $-1.333-4=\text{A}(-1.333)$ \\ ...
        } \\
        
    \end{tblr}
    \caption{Examples of the recognition errors in the ``Textual \&
Logical'' category in Table \ref{tab:error_taxonomy}. The incorrecrt recognitions are marked in \textcolor{red}{red} and the rectified versions are marked in \textcolor{green!60!black}{green}.}
    \label{tab:diagrammatic_error_logical}
\end{table}

To implement this module, we first aggregated all recognition errors flagged by human experts within the observation set; the process for collecting these errors is detailed in Section \ref{sec: crafting_gold_standard_recog_text}. Leveraging this prior knowledge, experts distilled a set of empirical criteria, presented in Figures \ref{prompt: recog err detection (Part I)} and \ref{prompt: recog err detection (Part II)}, to guide the detection of potential recognition errors in the test set. These criteria were then integrated into the recognition error detector, which identifies likely transcription failures in student responses that received score deductions in the initial grading stage. For the case study, we use \texttt{GPT-5.1} as the detector. Based on its output, the LLM regrader re-evaluates the affected student work using the prompt shown in Figure~\ref{prompt: regrading}. During this process, the regrader is explicitly instructed to disregard any content flagged as a recognition error, even if it introduces discrepancies with the reference solution.

In cases where the detector is unable to definitively distinguish between a recognition error and a genuine student mistake (i.e., segments marked as ``[uncertain]'' in Figure \ref{prompt: recog err detection (Part I)}), the corresponding solution is referred to a human grader for final evaluation. In our case study, this human grader is the doctoral teaching assistant for the course from which \dsname was derived.

\subsection{Human and MLLM API Settings}
\noindent\textbf{Experts and Human Graders} The graduate grader referenced in Table \ref{tab: exp_result1} is a graduate student who previously completed the course associated with \dsname and has also served as a teaching assistant for the class. All experts and human graders involved in the student work regrading are doctoral students with comprehensive mastery and deep subject-matter expertise in the course content.

\noindent\textbf{Model Configurations} All MLLMs used in our visual recognition experiments are accessed exclusively through their official API interfaces. To ensure reproducibility and reduce output variability, we set the temperature parameter to 0 for all models, except for \texttt{GPT-5.1}, whose temperature is fixed at 1 by the official setting. Unless otherwise specified, all other hyperparameters (e.g., reasoning effort for \texttt{GPT-5.1}) remain at their default API settings. Given the low-temperature configurations, which result in highly stable outputs, and the practical constraints of deployment, where inference is typically conducted only once to control costs, we report experimental results based on a single run per model.

\subsection{Examples about Recognition Error Taxonomy}
\label{sec: example_for_error_taxonomy}
In this section, we present several examples of recognition errors, categorized according to the taxonomy defined in Table \ref{tab:error_taxonomy}. The examples are organized as follows:

\begin{itemize}
\item \textbf{Symbolic \& Character:} Representative recognition errors in this category are provided in Table \ref{tab:symbol_error_examples}.
\item \textbf{Structural \& Notational:} Examples of errors involving equation structure or notation are shown in Table \ref{tab:eq_error_examples}.
\item \textbf{Diagrammatic:} Recognition errors related to diagrams are illustrated in Table \ref{tab:diagrammatic_error_examples}.
\item \textbf{Textual \& Logical:} Examples of misinterpretations involving textual descriptions or logical reasoning appear in Table \ref{tab:diagrammatic_error_logical}.
\end{itemize}

\clearpage

\begin{figure*}[htbp]
\begin{solutionbox}{Example 1 for data scope clarification (A recognized handwritten solution to Problem 9.8-2; involving foundations from physics, differential equations, and calculus)}

For $t < 0$:
\[ i(0) = \frac{-1}{1+4} = -0.2 \text{ A} \]
\[ v(0) = \frac{4}{1+4}(-1) = -0.8 \text{ V} \]

For $t > 0$:

KCL $\rightarrow$ $\frac{v-V_s}{R_1} + C \frac{dv}{dt} + i = 0$

KVL to right half $\rightarrow$ $v = R_2 i + L \frac{di}{dt}$

plug KVL eq. to KCL eq.:
\[ \frac{R_2 i + L \frac{di}{dt} - V_s}{R_1} + C \left[ R_2 \frac{di}{dt} + L \frac{d^2i}{dt^2} \right] + i = 0 \]
\[ \Rightarrow \frac{d^2i}{dt^2} + \frac{(L+R_1R_2C)}{R_1LC} \frac{di}{dt} + \frac{(R_1+R_2)}{R_1LC} i = \frac{V_s}{R_1LC} \]
\[ \Rightarrow \frac{d^2i}{dt^2} + 5\frac{di}{dt} + 5i = 1 \]

Forced response is constant, $i_f = B$ so $1 = \frac{d^2B}{dt^2} + 5\frac{dB}{dt} + 5B$

Therefore $B = 0.2 \text{ A}$

Characteristic eq: $s^2 + 5s + 5 = 0$

Roots $\Rightarrow s_1 = -3.62$, $s_2 = -1.38$

\[ i_n = A_1 e^{-3.62t} + A_2 e^{-1.38t} \]
\[ i(t) = A_1 e^{-3.62t} + A_2 e^{-1.38t} + 0.2 \]
\[ v(t) = \left(4 i(t) + 4 \frac{di(t)}{dt}\right) = -10.48 A_1 e^{-3.62t} - 1.52 A_2 e^{-1.38t} + 0.8 \]

At $t=0^+$,
\[ -0.2 = i(0^+) = A_1 + A_2 + 0.2 \Rightarrow A_1 + A_2 = -0.4 \]
\[ -0.8 = v(0^+) = -10.48 A_1 - 1.52 A_2 + 0.8 \Rightarrow -10.48 A_1 - 1.52 A_2 = -1.6 \]

Using matrix calc: $A_1 = 0.246$ and $A_2 = -0.646$
\[ i(t) = 0.246 e^{-3.62t} - 0.646 e^{-1.38t} + 0.2 \text{ A} \]

\end{solutionbox}
\caption{An example in our dataset including a student's solution on using foundations from physics, differential equations, and calculus to solve the target current.}
\label{fig: example_1_for_data_scope_clarificiation}
\end{figure*}

\begin{figure*}[htbp]
\begin{solutionbox}{Example 2 for data scope clarification (A recognized handwritten solution to Problem 10.6-4)}

Mesh 1:
$$ (40+j15)i_1 + (25-j50)(i_1-i_3) - 48 \angle 75^\circ = 0 $$
$$ (65-j35)i_1 - (25-j50)i_3 = 48 \angle 75^\circ $$

Mesh 2:
$$ 48 \angle 75^\circ + (-j50)(i_2-i_3) + (32+j16)i_2 = 0 $$
$$ (32-j34)i_2 + j50 i_3 = -48 \angle 75^\circ $$

Mesh 3:
$$ j40 i_3 - (-j50)(i_2-i_3) - (25-j50)(i_1-i_3) = 0 $$
$$ (-25+j50)i_1 + j50 i_2 + (25-j160)i_3 = 0 $$

$$
\begin{bmatrix}
-25+j50 & j50 & 25-j160 \\
0 & 32-j34 & j50 \\
65-j35 & 0 & -25+j50
\end{bmatrix}
\begin{bmatrix}
i_1 \\
i_2 \\
i_3
\end{bmatrix}
=
\begin{bmatrix}
0 \\
-48 \angle 75^\circ \\
48 \angle 75^\circ
\end{bmatrix}
$$

$i_1 = 0.794 \angle 111^\circ \text{ A}$, \quad
$i_2 = 0.790 \angle -61.7^\circ \text{ A}$, \quad
$i_3 = 0.229 \angle 176^\circ \text{ A}$

\end{solutionbox}
\caption{An example in our dataset including a student's solution on using linear algebra and complex operation to solve the target currents.}
\label{fig: example_2_for_data_scope_clarificiation}
\end{figure*}

\begin{figure*}[t] 
    \begin{promptbox}{Example: \texttt{Gemini-2.5-Pro} transcribed student handwritten solution in Figure \ref{fig: student example P9.4-2}}
> A circuit diagram for `t >= 0` is drawn, showing a capacitor, resistor, and inductor connected (*\colorbox{softred}{\ttfamily \detokenize{in a series}}*) (*\colorbox{softred}{\ttfamily \detokenize{loop}}*)  The capacitor is labeled `1/3 F` and `C`, with the voltage across it marked as `v(t)` (positive on the (*\colorbox{softred}{\ttfamily \detokenize{left}}*)). The resistor is labeled `R` and `3/4 \Omega`. The inductor is labeled `1H`. A clockwise current is indicated, with the label `i_L(t)` pointing downwards through the inductor.

**Circuit t>=0:**

**Var 1: v(t)**
$$C \frac{dv}{dt} + \frac{V(t)}{R} + i_L(t) = 0$$

**Var 2: i(t)**
$$i_c(t) = -C \frac{dv}{dt} - \frac{v(t)}{R}$$

$$L \frac{di_L}{dt} = v(t)$$

$$L(-C \frac{d^2v}{dt^2} - \frac{1}{R} \frac{dv}{dt}) = v(t)$$

$$-LC \frac{d^2v}{dt^2} - \frac{L}{R} \frac{dv}{dt} - v(t) = 0$$

$$- \frac{1}{3} v''(t) - \frac{4}{3} v'(t) - v(t) = 0$$

$$v(t) = c_1 e^{-t} + c_2 e^{-3t}$$

**initial conditions**
$$v(0)=2 = c_1 + c_2$$

(*\colorbox{softred}{\ttfamily \detokenize{$$v'(t) = c_1 e^{-t} - 3 c_2 e^{-3t}$$}}*)

(*\colorbox{softred}{\ttfamily \detokenize{$$v'(0) = c_1 - 3c_2 = 0$$}}*)

$$c_1 = 3$$

$$c_2 = -1$$

(*\colorbox{softred}{\ttfamily \detokenize{$$v(t) = [3e^{-t} - e^{-3t}]V + 20$$}}*)

> A graph is drawn with `v_c(t)` on the vertical axis and `t` on the horizontal axis. The vertical axis is labeled `V` and has a tick mark at `2`. The horizontal axis has tick marks at `2` and `4`. (*\colorbox{softred}{\ttfamily \detokenize{The plotted}}*) (*\colorbox{softred}{\ttfamily \detokenize{curve starts at `(0, 2)`, increases to a peak value at approximately `t=0.5`, and then decays}}*), approaching zero.
    \end{promptbox}
\caption{Transcription generated by \texttt{Gemini-2.5-Pro} from the student handwritten solution shown in Figure \ref{fig: student example P9.4-2}. Expert-identified recognition errors are highlighted in \colorbox{softred}{red}.}
\label{prompt: student example P9.4-2 gemini recog}
\end{figure*}

\begin{figure*}[t] 
    \begin{promptbox}{Example: Expert-rectified \texttt{Gemini-2.5-Pro} transcribed student handwritten solution in Figure \ref{fig: student example P9.4-2}}
> A circuit diagram for `t >= 0` is drawn, showing a capacitor, resistor, and inductor connected (*\colorbox{green!20}{\ttfamily \detokenize{in parallel}}*). The capacitor is labeled `1/3 F` and `C`, with the voltage across it marked as `v(t)` (positive on the (*\colorbox{green!20}{\ttfamily \detokenize{top}}*)). The resistor is labeled `R` and `3/4 \Omega`. The inductor is labeled `1H`. A clockwise current is indicated, with the label `i_L(t)` pointing downwards through the inductor.

**Circuit t>=0:**

**Var 1: v(t)**
$$C \frac{dv}{dt} + \frac{V(t)}{R} + i_L(t) = 0$$

**Var 2: i(t)**
$$i_c(t) = -C \frac{dv}{dt} - \frac{v(t)}{R}$$

$$L \frac{di_L}{dt} = v(t)$$

$$L(-C \frac{d^2v}{dt^2} - \frac{1}{R} \frac{dv}{dt}) = v(t)$$

$$-LC \frac{d^2v}{dt^2} - \frac{L}{R} \frac{dv}{dt} - v(t) = 0$$

$$- \frac{1}{3} v''(t) - \frac{4}{3} v'(t) - v(t) = 0$$

$$v(t) = c_1 e^{-t} + c_2 e^{-3t}$$

**initial conditions**
$$v(0)=2 = c_1 + c_2$$

(*\colorbox{green!20}{\ttfamily \detokenize{$$v'(t) = -c_1 e^{-t} - 3 c_2 e^{-3t}$$}}*)

(*\colorbox{green!20}{\ttfamily \detokenize{$$v'(0) = -c_1 - 3c_2 = 0$$}}*)

$$c_1 = 3$$

$$c_2 = -1$$

(*\colorbox{green!20}{\ttfamily \detokenize{$$v(t) = [3e^{-t} - e^{-3t}]V , t \ge 0$$}}*)

> A graph is drawn with `v_c(t)` on the vertical axis and `t` on the horizontal axis. The vertical axis is labeled `V` and has a tick mark at `2`. The horizontal axis has tick marks at `2` and `4`. (*\colorbox{green!20}{\ttfamily \detokenize{The plotted}}*)(*\colorbox{green!20}{\ttfamily \detokenize{curve starts from negative `t`, increases to a peak value `2` at `t=0`, and then decays}}*), approaching zero.
    \end{promptbox}
\caption{Expert-rectified version of the \texttt{Gemini-2.5-Pro} transcription from the student handwritten solution shown in Figure \ref{fig: student example P9.4-2}. Corrections corresponding to the original recognition errors are highlighted in \colorbox{green!20}{green}.}
\label{prompt: student example P9.4-2 expert recog}
\end{figure*}

\begin{figure*}[t] 
    \begin{promptbox}{Example: \texttt{GPT-5.1} transcribed student handwritten solution in Figure \ref{fig: student example P8.3-10}}
> [On the left, the student redraws the original circuit but with the inductor removed (*\colorbox{softred}{(open-circuited)}*) to find the initial voltage across where the inductor will be. From left to right there is a 20 V source in series with a 4 \Omega resistor, followed by a node that branches downward to a 12 \Omega resistor, then to a 9 A current source, then to a 3 \Omega resistor, all in parallel. To the right of this parallel network is a 9 \Omega resistor whose right terminal is labeled with $+$ at the top and $-$ at the bottom as $v$.  
> The student then replaces the 4 \Omega, 12 \Omega, and 3 \Omega resistors and the 9 A current source with equivalent sources/resistances: first, turning the 20 V source and 4 \Omega resistor into an equivalent 5 A current source in parallel with 4 \Omega, then showing that 5 A source in parallel with the 12 \Omega resistor, the 9 A source, and the 3 \Omega resistor, all feeding the 9 \Omega resistor where $v$ is measured.  
> Next, the two current sources (5 A and 9 A) are added in parallel to yield a single 14 A source in parallel with an equivalent resistance made from the 4 \Omega, 12 \Omega, and 3 \Omega resistors in parallel.  
> Finally, these parallel resistors are combined into (*\colorbox{softred}{\ttfamily \detokenize{a single 3 \Omega resistor}}*) in series with the 9 \Omega resistor to the right, still labeled $v$ with $+$ at the top and $-$ at the bottom.  
> On the right side of the page, the student draws the circuit for $t>0$ where the inductor is connected: a 3 \Omega resistor in series with a 9 \Omega resistor, and a $1/2$ H inductor in series with them, with the voltage across the 9 \Omega resistor labeled $v$.]

---

**Left side (initial conditions and equivalent circuit)**

$$\text{(top node currents)} \Rightarrow 5\text{A} \quad 4\Omega \quad 12\Omega \quad 9\text{A} \quad 3\Omega \quad 9\Omega \; v^+$$

*add curr srcs in parallel*

$$14\text{A} \quad \boxed{\frac{3}{2}\,\Omega} \quad 9\Omega^{+}_{-}$$

$$\boxed{3\Omega} \quad 9\Omega^{+}_{-}$$

$$\frac{1}{\frac{1}{4} + \frac{1}{12} + \frac{1}{3}} \Rightarrow \frac{12}{8}$$

$$V = 9\left(\frac{21}{9 + \frac{3}{2}}\right)\,V$$

$$V = \frac{189}{\frac{21}{2}}\,V$$

$$V = 18\,V$$

---

**Right side (differential equation and solution)**

> [Small circuit: a 3 \Omega resistor in series with a $1/2$ H inductor and a 9 \Omega resistor, with $+$ at the left of the 9 \Omega and $-$ at the right labeled $v$.]

(*\colorbox{softred}{\ttfamily \detokenize{$$0 = 12\,i_L(t) + \tfrac{1}{2}\frac{d}{dt} i(t)$$}}*)

(*\colorbox{softred}{\ttfamily \detokenize{$$i_L'(t) = C e^{-24t}$$}}*)

(*\colorbox{softred}{\ttfamily \detokenize{$$i_{L,c}(t) = B \quad \Rightarrow 0 = 24 B$$}}*)

$$i_L(t) = C e^{-24t}$$

$$V(t) = 9 C e^{-24t}$$

$$V(0) = 18 = 9 C e^{0}$$

$$C = 2$$

(*\colorbox{softred}{\ttfamily \detokenize{$$\boxed{V(t) = 18 e^{-24t},\quad t>0}$$}}*)
    \end{promptbox}
\caption{Transcription generated by \texttt{GPT-5.1} from the student handwritten solution shown in Figure \ref{fig: student example P8.3-10}. Expert-identified recognition errors are highlighted in \colorbox{softred}{red}.}
\label{prompt: student example P8.3-10 GPT recog}
\end{figure*}

\begin{figure*}[t] 
    \begin{promptbox}{Example: Expert-rectified \texttt{Gemini-2.5-Pro} transcribed student handwritten solution in Figure \ref{fig: student example P8.3-10}}
> [A sequence of circuit diagrams shows the simplification of the circuit for t < 0 to find an initial condition. The inductor is omitted and the switch is treated as a (*\colorbox{green!20}{\ttfamily \detokenize{short circuit}}*).]
>
> 1.  **Initial Diagram:** A redrawing of the original circuit components relevant for t < 0. A 20V source is in series with a 4\Omega resistor. This is in parallel with a 12\Omega resistor, a 9A current source, a 3\Omega resistor, and a 9\Omega resistor. The voltage `v` is marked across the 9\Omega resistor.
> 2.  **Source Transformation:** `=>` The 20V source and 4\Omega series resistor are transformed into a 5A current source in parallel with a 4\Omega resistor.
> 3.  **Combine Parallel Elements:** `=>` The parallel current sources (5A and 9A) are combined into a single 14A source. The parallel resistors (4\Omega, 12\Omega, and 3\Omega) are combined into (*\colorbox{green!20}{\ttfamily \detokenize{a single `3/2 \Omega` resistor}}*). The circuit is now a 14A source in parallel with a `3/2 \Omega` resistor, connected to the 9\Omega resistor.
> 4.  **Final Source Transformation:** `=>` The 14A source and `3/2 \Omega` parallel resistor are transformed into a 21V voltage source in series with a `3/2 \Omega` resistor. This is connected to the 9\Omega resistor.

$$
\begin{aligned}
v &= 9 \left( \frac{21}{9 + \frac{3}{2}} \right) v \\
v &= 189 \left( \frac{2}{21} \right) v \\
v &= 18V
\end{aligned}
$$

---
> [A circuit diagram for t > 0 is drawn. It shows a 3\Omega resistor, a 9\Omega resistor, and a 1/2 H inductor connected in a series loop. The voltage `v` is labeled across the 9\Omega resistor with the positive terminal on the left.]

(*\colorbox{green!20}{\ttfamily \detokenize{$$0 = 12 i_L(t) + \frac{1}{2} \frac{di_L}{dt}(t)$$}}*)

(*\colorbox{green!20}{\ttfamily \detokenize{$$i_{Ln}(t) = C e^{-24t}$$}}*)

(*\colorbox{green!20}{\ttfamily \detokenize{$$i_{L_F}(t) = B \rightarrow 0 = 24B$$}}*)

$$i_L(t) = C e^{-24t}$$

$$V(t) = 9 C e^{-24t}$$

$$V(0) = 18 = 9 C e^0$$

$$C = 2$$

(*\colorbox{green!20}{\ttfamily \detokenize{$$\boxed{V(t) = 18e^{-24t} V, \quad t > 0s}$$}}*)
    \end{promptbox}
\caption{Expert-rectified transcription from \texttt{Gemini-2.5-Pro} for the student handwritten solution shown in Figure \ref{fig: student example P8.3-10}. The highlighted segments in \colorbox{green!20}{green} correspond to parts where \texttt{GPT-5.1} (used here as the target model for diversity in few-shot learning) produced recognition errors, as shown in Figure \ref{prompt: student example P8.3-10 GPT recog}. The LLM-based error detector is expected to produce a recognition error report similar to that shown in Figure \ref{prompt: example_about_recognition_error} when comparing these transcriptions. Note that in practice some MLLMs may transcribe students' sketches while others may not, we selected to be tolerate to their differences (especially to those not clearly recorded in the expert-rectified transcriptions).}
\label{prompt: student example P8.3-10 expert recog}
\end{figure*}

\begin{figure*}[t] 
    \begin{promptbox}{Prompt for student handwritten solution transcription}
You are an expert OCR (Optical Character Recognition) assistant specializing in student-submitted work for engineering and physics problems.
Your primary goal is to accurately transcribe the content from an image of a student's handwritten work, including text, mathematical formulas, and diagrams.
To provide you with the necessary context to improve accuracy, I am giving you the original problem statement (text and any associated images) that the student was working on.

---

## 1. Context: The Original Problem
This is the problem the student was asked to solve. Use this information to better interpret their handwriting, especially for technical terms, symbols, numbers, and to compare against any diagrams they draw.

**Original Problem Statement Text:**
{problem_statement}

**Original Problem Images:**
(If any images are associated with the original problem, they will be provided as input right after this prompt.)

---

## 2. Your Task: Transcribe the Student's Work
Now, analyze the **final image provided**, which contains the **student's handwritten work**.

**Instructions:**
1.  **Identify the Core Content**: Focus on the student's actual work (the text, equations, diagrams, and code) in their image.

2.  **Analyze and Describe Circuit Diagrams**: This is a critical step. Students often redraw circuits to aid their analysis. You must analyze any circuit diagrams drawn by the student and describe them **only if they add new information** compared to the original problem's diagram.
    *   **A. For Annotations:** If the student redraws the circuit and adds labels like node names (`A`, `B`, `v_1`), current arrows (`i_x`, `i_1`), or voltage polarities (`+ V_o -`), describe these additions in natural language. Frame your description inside a Markdown blockquote.
        > **Example Output for Annotations:**
        > `[A circuit diagram is drawn with a current source, a voltage source, and a resistor in series. The current source is labeled 2A. The voltage source is labeled 10V. The resistor is labeled 5\Omega, and the voltage across it is labeled V1 with the positive terminal on the left.]`

    *   **B. For Simplifications or Transformations:** If the student shows a sequence of diagrams representing circuit simplification (e.g., combining resistors, source transformation, finding Thévenin/Norton equivalents), describe this process step-by-step. Use arrows `=>` to show the progression. Frame your description inside a Markdown blockquote.
        > **Example Output for Simplifications:**
        > `[Circuit with the two rightmost 1\Omega resistors in series is combined into a single 2\Omega resistor] => [The new 2\Omega resistor is now in parallel with the existing 2\Omega resistor, which are combined into a 1\Omega equivalent resistor] => [Final equivalent circuit with a 12V source and a 3\Omega resistor]`

3.  **Accurate Text Transcription**: Transcribe all other handwritten text as accurately as possible. For code, preserve indentation and syntax.

4.  **Mathematical Formula Formatting**:
    *   Convert all mathematical expressions, equations, and formulas to proper LaTeX syntax.
    *   Use inline math with single dollar signs: `$formula$` for simple expressions.
    *   Use display math with double dollar signs: `$$formula$$` for equations that should be centered on their own line.
    *   Ensure all mathematical symbols, subscripts, superscripts, fractions, integrals, derivatives, etc. are properly formatted in LaTeX.
    *   Examples: `$x^2 + y^2 = z^2$`, `$$\int_0^1 f(x) dx$$`, `$\frac{a}{b}$`, `$\sum_{i=1}^n x_i$`

5.  **Formatting**: Use appropriate Markdown formatting (headings, lists, bold text) for clarity while preserving the student's original organization.

6.  **Output**: Return only the final Markdown content of the student's transcribed work. Do not include any introductory phrases or explanations about the transcription process.
    \end{promptbox}
\caption{Prompt used to instruct \texttt{Gemini-3-Pro-Preview}, \texttt{Gemini-2.5-Pro}, \texttt{GPT-5.1}, and \texttt{Claude-4.5-Sonnet} to transcribe student handwritten solutions into Markdown format.}
\label{prompt: transcription_common}
\end{figure*}

\begin{figure*}[t] 
    \begin{promptbox}{Prompt for student handwritten solution transcription (for \texttt{Qwen3-VL} series)}
You are a **Strict Verbatim OCR Transcription System** specializing in digitizing handwritten student work for engineering and physics.
Your **SOLE** purpose is to convert the pixels of the student's handwriting into digital text and LaTeX.
You are **NOT** a tutor or a solver. Do not attempt to solve the problem, correct the student's math, or fill in missing steps.

To provide you with context for ambiguous handwriting, I am giving you the original problem statement.

---

## 1. Context: The Original Problem
(Use this ONLY to recognize messy handwriting. **DO NOT** transcribe or describe this content as the student's work.)

**Original Problem Statement Text:**
{problem_statement}

**Original Problem Images:**
(Provided as input if available)

---

## 2. Your Task: Transcribe the Student's Work
Now, analyze the **final image provided**, which contains the **student's handwritten work**.

**CRITICAL RULES (Anti-Hallucination):**
*   **NO AUTO-COMPLETION**: If the student wrote `F = m`, do NOT output `F = ma`. Output exactly `F = m`.
*   **NO INFERENCE**: Do not generate intermediate mathematical steps that are not explicitly written on the paper.
*   **NO GHOST DIAGRAMS**: If the student did not draw a circuit diagram with their pen/pencil, **do not output any diagram description**. Do not describe the printed diagrams from the original problem.
*   **PRESERVE ERRORS**: If the student made a mistake, transcribe the mistake exactly as written.

**Instructions:**

1.  **Identify the Student's Ink**: Ignore printed text from the worksheet/exam paper unless the student has written over it. Focus only on the handwritten content.

2.  **Analyze Student-Drawn Diagrams (STRICTLY CONDITIONAL)**:
    *   **Check First**: Did the student actually **redraw** or **annotate** a circuit diagram by hand?
    *   **If NO**: Skip this step entirely. Do not describe the printed problem image.
    *   **If YES (and only if new information is added)**:
        *   **For Annotations**: If the student added labels or arrows to a drawn circuit, describe them inside a Markdown blockquote.
            > **Example:** `[Student's hand-drawn circuit: Labeled node A and B, added current arrow i_x pointing right.]`
        *   **For Simplifications**: If the student drew a sequence of simplified circuits, describe the progression using arrows `=>`.
            > **Example:** `[Hand-drawn diagram showing 2\Omega and 3\Omega resistors combined] => [Final simplified circuit with source transformation]`

3.  **Strict Text Transcription**: Transcribe handwritten text exactly as it appears.

4.  **Mathematical Formula Formatting**:
    *   Convert handwritten math to proper LaTeX.
    *   Use inline math `$...$` and display math `$$...$$` appropriately.
    *   **Verify against the image**: Ensure every symbol in your LaTeX output exists in the student's handwriting.

5.  **Output Format**: Return **only** the Markdown content of the transcription. Start directly with the content.
    \end{promptbox}
\caption{Prompt used to instruct \texttt{Qwen3-VL-Plus} and \texttt{Qwen3-VL-8B-Thinking} to transcribe student handwritten solutions into Markdown format.}
\label{prompt: transcription_qwen}
\end{figure*}

\begin{figure*}[t] 
    \begin{promptbox}{One-shot prompt for the LLM detector to find recognition errors in the MLLM recognized text}
You are an expert OCR (Optical Character Recognition) Quality Assurance Judge.
Your task is to compare a [TARGET MODEL OUTPUT] against a verified [GROUND TRUTH].

### GENERAL COMPARISON RULES:
1. **IGNORE Syntax Differences:** Do not report benign LaTeX variations (e.g., `\\frac` vs `\\dfrac`, extra whitespaces) if the rendered mathematical meaning is identical.
2. **IGNORE Minor Wording:** Do not report missing conversational filler words (e.g., "So," "Therefore") unless they change the logic. Ignore case differences (e.g., $v$ vs $V$).
3. **ABOUT Redrawn Diagrams:** If the OCR result depicts a student's redrawn diagram, ignore layout differences unless the values, units, or directions (e.g., 2\Omega vs 3\Omega, clockwise vs counterclockwise) are clearly incorrect compared to the Ground Truth logic.

### OUTPUT FORMAT:
If errors are found, list them in the following structured format:
1. 
Source: [Exact snippet from TARGET that is wrong]
Rectified Version: [Correct snippet from GROUND TRUTH]
Reason: [Brief explanation of the error]

If no meaningful errors are found, output: "No significant errors found."

### ONE-SHOT EXAMPLE:
**[EXAMPLE TARGET]:**
{examples['target']}

**[EXAMPLE GROUND TRUTH]:**
{examples['gt']}

**[EXAMPLE EXPECTED JUDGEMENT]:**
{examples['result']}

**[EXAMPLE CORRECTION LOGIC]:**
{examples['evidence']}

---

### CURRENT TASK:

**[TARGET MODEL OUTPUT]:**
{target_content}

**[GROUND TRUTH]:**
{label_content}

**[CAUTION]:**
You should report the differences in the formula formatting if they break the formula's mathematical meaning, e.g., "\int tdt" vs "\int t\tau" (the dummy variable changes from t to \tau will lead to a different numerical result), "di/dt + i = 0" vs "di/dt + i_L = 0", etc.

Again, you can ignore the following differences:
1. The differences in the crossed-out or cancelled content, unless only one of the ground truth/target's content is crossed-out or cancelled.
2. Capitalization differences (e.g., $v$ vs $V$, $v_0$ vs $V_o$, $w$ and $W$) **unless** they break the formula's mathematical meaning (e.g., both $v$ vs $V$ expressing the same physical quantity in a single formula like $dv/dt + V = 0$, "\int tdt" vs "\intt\tau", etc.).
3.  Formating or presenting differences which won't change the formula's mathematical meaning/result (e.g., $j2$ vs $2j$, $3A$ vs $3(B-2)$ where it's written that A=B-2 before, "5e^7" vs "5 e^7.0", "100mH vs 0.1H", "A = 3" vs "A -> 3", "cos 2t" vs "cos(2t)", -$\frac{1}{2/3}$ vs $\frac{-3}{2}$, etc.) should be ignored. They make the formula/sentence looks different, but they are mathematically/numerically equivalent.

**Your Judgement:**
    \end{promptbox}
\caption{Prompt used by the LLM-enabled recognition error detector to identify potential recognition errors in the target transcription (i.e., \texttt{target\_content}) by comparing it with the expert-verified gold standard (i.e., \texttt{label\_content}). To enhance detection performance, a one-shot learning approach is employed by providing an example that includes expert-annotated recognition errors and their corresponding rectifications.}
\label{prompt: recognition effect comparison}
\end{figure*}

\begin{figure*}[t] 
    \begin{promptbox}{One-shot example for LLM recognition error detector: errors and their rectified versions}
1.
Source:
On the left, the student redraws the original circuit but with the inductor removed \colorbox{softred}{(open-circuited)} to find the initial voltage across where the inductor will be.

Rectified Version:
On the left, the student redraws the original circuit but with the inductor removed (short-circuited) to find the initial voltage across where the inductor will be.

2
Source:
> Finally, these parallel resistors are combined into a single 3 \Omega resistor in series with the 9 \Omega resistor to the right, still labeled $v$ with $+$ at the top and $-$ at the bottom.

Rectified Version:
> Finally, these parallel resistors are combined into a single 3/2 \Omega resistor in series with the 9 \Omega resistor to the right, still labeled $v$ with $+$ at the top and $-$ at the bottom.

3.
Source:
$$
0 = 12\,i_L(t) + \tfrac{1}{2}\frac{d}{dt} i(t)
$$

Rectified Version:
$$
0 = 12\,i_L(t) + \tfrac{1}{2}\frac{d}{dt} i_L(t)
$$

4.
Source:
$$
i_L'(t) = C e^{-24t}
$$

Rectified Version:
$$
i_{Ln}(t) = C e^{-24t}
$$

5.
Source:
$$
i_{L,c}(t) = B \quad \Rightarrow 0 = 24 B
$$

Rectified Version:
$$
i_{Lf}(t) = B \quad \Rightarrow 0 = 24 B
$$

6.
Source:
\boxed{V(t) = 18 e^{-24t},\quad t>0}

Rectified Version:
\boxed{V(t) = 18 e^{-24t},\quad t>0 s}
    \end{promptbox}
\caption{Recognition errors and their corresponding rectifications for the sample shown in Figure \ref{fig: student example P8.3-10}, based on the comparison between the MLLM-generated transcription and the expert-verified gold standard. In this example, six item-level recognition errors were identified and corrected. The output format of the LLM-based recognition error detector follows this same structure.}
\label{prompt: example_about_recognition_error}
\end{figure*}

\begin{figure*}[t]
    \begin{promptbox}{One-shot example for LLM recognition error detector: explanations for capturing some errors}
Items 1 and 2 represent entirely different scenarios for the target diagram (e.g., "open-circuited" vs. "short-circuited", or "3 \Omega resistor in series with 9 \Omega" vs. "3/2 \Omega resistor in series with 9 \Omega"). These must be flagged as errors. Other variations such as minor grammatical differences or descriptions that are slightly more or less detailed can be ignored as long as they do not convey incorrect information.

In Item 3, changing $i_L(t)$ to $i(t)$ leads to a misunderstanding of the function; symbols representing the same physical quantity must remain consistent throughout the formula. Similarly, errors that will significantly change the formulas meaning must be pointed out, for example, if the ground truth uses $q = \int_{-\infty}^{t} i(\tau) \, d\tau$ but the target uses $dt$ instead of $d\tau$, this totally changes the mathematical meaning. Nevertheless, it is acceptable to ignore differences that do not change the underlying meaning, such as $v_0$ vs. $V_o$ (capitalization), $10^{-7}$ vs. $E^{-7}$, $\int_{0}^{1}d\tau$ vs. $\int_{0}^{1}dx$, $10^\circ$ vs. $10 \text{ degrees}$, $sin 2t$ vs $sin(2t)$, $\frac{1}{3/2}$ vs. $\frac{2}{3}$, etc. These are simply diverse ways of expression that lead to the same result without causing confusion.

Items 4 and 5 represent standard errors. However, note that errors involving symbols for widely used physical quantities (such as natural response and forced response) must be treated strictly. For instance, $I_n$ should not be misidentified as $I_e, I_x$, or $I_t$; $V_f$ should not be $V_{eq}$ or $V_h$; the steady-state value $k$ or $K$ should not be altered; and Thevenin resistance ($R_t, R_T, R_{th}$) should not be labeled as $R_x$.

Item 6 involves unit errors. It is acceptable to be lenient with unit issues in intermediate steps; however, for final answers, any missing units or the addition of incorrect units (those that do not match the ground truth) must be treated as errors.

The **sentence order** (e.g., "first A then B" vs. "first B then A") and **layout sketches** in the OCR results, such as:

"
---

**Left side (initial conditions and equivalent circuit)**

$$
\text{(top node currents)} \Rightarrow 5\text{A} \quad 4\Omega \quad 12\Omega \quad 9\text{A} \quad 3\Omega \quad 9\Omega \; v^+
$$

*add curr srcs in parallel*

$$
14\text{A} \quad \boxed{\frac{3}{2}\,\Omega} \quad 9\Omega^{+}_{-}
$$

$$
\boxed{3\Omega} \quad 9\Omega^{+}_{-}
$$

$$
\frac{1}{\frac{1}{4} + \frac{1}{12} + \frac{1}{3}} \Rightarrow \frac{12}{8}
$$

"

can be ignored. Focus solely on information present in the ground truth that the OCR results lack (excluding sketch-like elements), especially missing critical variables, key derivation steps, and algebraic expressions used for substitution.

Formulas may be presented in different algebraic forms, such as:

$$
V = \frac{189}{\frac{21}{2}}\,V
$$

and 

$$
v = 189 \left( \frac{2}{21} \right) v
$$

These should be treated as identical. 
    \end{promptbox}
\caption{Explanations for each item-level recognition error presented in Figure \ref{prompt: example_about_recognition_error}. These justifications serve as reference examples to guide the LLM-based detector in identifying and interpreting recognition errors.}
\label{prompt: explanation_for_example_about_recognition_error}
\end{figure*}

\begin{figure*}[t]
    \begin{promptbox}{A rubric example for question P9.7-2 in \dsname}
### M - Method
Uses an appropriate method to find the forced (particular) response for v_f for each input (constant, ramp, exponential). Acceptable methods include time-domain particular-solution ansatz, or Laplace-domain solution; should clearly target the forced response (not the homogeneous/natural response).
- **Typical deduction**: 0.01
- **Max deduction**: 0.04
**Deduction rules:**
- **M1**: Solves only the homogeneous/natural response or mixes total response with initial conditions instead of isolating the forced response. - deduct 0.01. _e.g.,_ Gives v(t)=C1 e^{s1 t}+C2 e^{s2 t} only; Uses ICs to determine constants for natural response but no particular solution
- **M2**: Chooses an incompatible particular form for the input type (e.g., uses sinusoidal trial for a ramp or exponential input). - deduct 0.01. _e.g.,_ Assumes v_f=Ae^{jwt} for vs=0.2 t; Assumes v_f=K (constant) for vs=e^{-30 t}

    
### C - Calculation
Carries out algebra, differentiation, and numeric substitution correctly (e.g., R/L=70 s^-1, 1/LC=12000 s^-2) and solves for constants in the particular solution accurately.
- **Typical deduction**: 0.01
- **Max deduction**: 0.04
**Deduction rules:**
- **C1**: Arithmetic or algebra mistakes in solving for particular-solution constants (e.g., A, B) after correct setup. - deduct 0.01. _e.g.,_ Incorrect A from 2400=12000A; Wrong B from 70A+12000B=0
- **C2**: Differentiation errors for the chosen trial (wrong sign or power). - deduct 0.01. _e.g.,_ d/dt(e^{-30t})=+30e^{-30t}
- **C3**: Numerical parameter errors (e.g., wrong 1/LC or R/L) that affect the final constants. - deduct 0.01. _e.g.,_ Uses 1/LC=1200 instead of 12000

### E - Equation
Sets up correct governing equations (KVL and i=C dv/dt) and the correct linear ODE relating v and v_s, with proper coefficients (R/L, 1/LC), and applies the chosen particular form correctly in the equation.
- **Typical deduction**: 0.01
- **Max deduction**: 0.04
**Deduction rules:**
- **E1**: Incorrect or incomplete differential equation relating v and v_s for the given RLC (e.g., wrong coefficients, sign errors, or omits a term). - deduct 0.01. _e.g.,_ Uses R/C instead of R/L; Misses the 1/LCv term
- **E2**: Misapplies element laws in equations (e.g., i!=C dv/dt or KVL written inconsistently with polarity). - deduct 0.01. _e.g.,_ Writes i= (1/C) dv/dt; Drops the inductor voltage term in KVL
- **E3**: Does not correctly substitute the trial particular form into the ODE for the given input. - deduct 0.01. _e.g.,_ Uses v_f=At+B but substitutes derivatives inconsistently

### U - Units
Includes correct units for final voltage answers and consistent time units. Voltages should be labeled in V; time assumed in seconds unless otherwise stated.
- **Typical deduction**: 0.01
- **Max deduction**: 0.02
**Deduction rules:**
- **U1**: Missing or incorrect units on final v_f(t) answers. - deduct 0.01. _e.g.,_ Reports v_f(t)=... without 'V'
- **U2**: Inconsistent time units that contradict given data (e.g., mixing ms and s without conversion). - deduct 0.01.

### NC - Not complete
Solution is missing required parts (a), (b), or (c), or does not present the final forced response expression(s) clearly.
- **Typical deduction**: 0.05
- **Max deduction**: 0.2
**Deduction rules:**
- **NC1**: Omits the forced response for any one of the three source cases. - deduct 0.05. _e.g.,_ Only provides (a) and (b)
- **NC2**: Shows work but does not state the final v_f(t) expression(s). - deduct 0.05.
- **NC3**: Provides only intermediate transforms/ODE without returning to time-domain forced response. - deduct 0.05.

## Notes
Full credit solutions typically: derive or state the correct second-order ODE for v and v_s for the given RLC (or an equivalent Laplace-domain relation), choose an appropriate particular form for each input (constant: v_f=K; ramp: v_f=At+B; exponential: v_f=Ae^{-30t}), solve for constants correctly using the circuit parameters (R=7 \Omega, L=0.1 H, C=833.3 \muF so R/L=70 s^-1 and 1/LC=12000 s^-2), and report v_f(t) with units of volts. Alternative correct methods (time-domain or Laplace) are acceptable as long as they target the forced solution.
    \end{promptbox}
\caption{Grading rubric for question P9.7-2, outlining how to identify student mistakes and assign score deductions across five categories: ``E'', ``M'', ``C'', ``U'', and ``NC'', as defined in Table \ref{tab:rubric}.}
\label{prompt: rubric example}
\end{figure*}

\begin{figure*}[t]
    \begin{promptboxsmallfont}{Prompt (Round 1) for LLM grading (generate deducted scores if any)}
You are a meticulous TA grader.
You will receive: (1) the problem statement (+ optional images), (2) a student's solution in Markdown recognized from their handwriting, and (3) a compact rubric snippet.
Your job: identify which rubric criteria apply and produce strictly valid JSON with per-criterion deductions and short justifications.

Rules:
- Judge only using the provided materials. If the recognized text seems partially wrong or noisy, rely on clear evidence.
- Use the rubric snippet's rules (codes M, E, C, U, NC). If multiple rules under the same code apply, you may include multiple items for that code.
- Do not exceed the rubric's typical intentions; avoid double-counting the same mistake under different rules unless they are clearly distinct.
- Use conservative rounding and keep messages concise.
- Output strictly valid JSON matching the schema provided below. No extra keys, no Markdown fences.

Schema (you MUST follow this exactly):
{
  "per_criterion": {
    "M": {
      "applied": [ {"rule_id": "string or empty", "deduct": number, "comment": "short justification"} ],
      "subtotal": number
    },
    "E": { "applied": [ ... ], "subtotal": number },
    "C": { "applied": [ ... ], "subtotal": number },
    "U": { "applied": [ ... ], "subtotal": number },
    "NC": { "applied": [ ... ], "subtotal": number }
  },
  "overall_comment": "1-3 sentences summary"
}

Notes:
- 'rule_id' can be empty if you cannot map to a specific micro-rule, but prefer using it when obvious from rubric snippet.
- 'subtotal' is the model's suggested deduction for that code BEFORE any external capping.
- Do not include total score; it will be post-processed with rubric caps externally.
"""

SCORING_USER_TEMPLATE = """PROBLEM STATEMENT:
{problem_statement}

OFFICIAL PROBLEM SOLUTION (ground truth, optional):
{final_answer}

PROBLEM IMAGES (optional base64 PNGs): count={num_images}
{images_hint}

STUDENT RECOGNIZED SOLUTION (Markdown):
{student_markdown}

RUBRIC SNIPPET (compact):
{rubric_snippet}

IMPORTANT:
- Apply deductions only if clearly supported by the content in the student solution.
- When students don't add unit for calculated values, you should not deduct for that criterion "U" (such as $V(t)$, $I(t)$, $R$, $L$, $C$, $\tau$, $Z_L$, $Z_C$, etc.). Only deducting when students clearly provide the wrong unit (such as $V(t)$ is given in $A$).
- If the student uses some values that are not stated in the problem solving process, try to find some clues from the PROBLEM IMAGES (such as the current $i_2$ labeled in the schematic). If you cannot find any clues, then you can deduct for that criterion "NC".
- If evidence is insufficient, set zero deductions for that criterion.
- It is fine for students not to simply a value (such as not turning 150/350 into 3/7) if the value is correct numerically or to simply a value (such as turn 4000/3998 into 1) when they are very close. 
- It is fine for students to use different precision's answer (e.g., 3.33, 3.333, 3.3, 10/3) or different units (e.g., mA vs A, mH vs H, \mu F vs F once they can be converted to the same value) for the same problem (e.g., 1.33 mA instead of 1.333 mA). Importantly, the rounding errors within a reasonable range should not be considered as calculation errors.
- You should recognize the equivalences between different number formats (e.g., decimal, fraction, scientific notation) and between different values under unit conversions when comparing student answers with the final answer or checking intermediate steps. 
- Even though the student use different types of KCL/KVL equations to solve the problem (such as 'sum of currents leaving = 0' and 'in=out'), which might looks inconsistent, you should not deduct for that criterion "E".
- The circuit diagram descriptions in the student solution may also contain important information. Thus, the circuit diagram descriptions should also be considered as part of the solution. 
- Fill all five keys M,E,C,U,NC in "per_criterion" (empty 'applied' list and subtotal=0 if no issues).
- If there is any deleted or crossed-out content in the student's Markdown, you should NOT consider it as part of the solution.
- If there is some content in a box, you should consider it as part of the solution and carefully check it.
- For the problems involving passive sign convention, if the final numerical answer is correct but the student did not make the passive sign convention explicit in their solution, you should NOT deduct any points.
- If the method or intermediate equations used by the student is different from the one in the reference solution but is valid and leads to correct results, you should NOT deduct any points.

Return ONLY JSON per the schema.
    \end{promptboxsmallfont}
\caption{First-round prompt used by the LLM grader to generate point deductions based on detected mistakes in the student’s solution.}
\label{prompt: 1st round grading}
\end{figure*}

\begin{figure*}[t]
    \begin{promptbox}{Prompt (Round 2) for LLM grading (explaining the score deductions)}
You are a careful grading explainer.
You will receive:
- The original student solution (Markdown),
- The rubric snippet,
- And the model's raw scoring JSON from the prior turn.

Your job:
1) For EVERY applied deduction (any code M/E/C/U/NC with subtotal > 0 or non-empty 'applied'), explain *why* with short, specific reasons tied to the student's content.
2) Show a concrete EVIDENCE SNIPPET: an exact, short substring copied from the student's Markdown that motivated the deduction (when possible).
3) Provide a single OVERALL CONFIDENCE label: "high" or "low". Use "low" when OCR noise/ambiguity/insufficient evidence could materially affect the scoring; otherwise "high".
4) Produce an EDITED MARKDOWN where ONLY the problematic parts are visually flagged **in-place**, keeping all other text untouched:
   - Wrap error spans with: <span style="background-color:#ffd6d6;color:#b00020;font-weight:bold"> ... </span>
   - Immediately after the span, append a small bracket note like: [deduct -0.03, reason: ...]
   - Do not move or rewrite surrounding text; preserve original formatting as much as possible.

Return strictly valid JSON with this schema:
{
  "reasons": {
    "M": [ {"rule_id": "string or empty", "deduct": number, "evidence_snippet": "short exact copy from student's text", "why": "short reason"} ],
    "E": [ ... ],
    "C": [ ... ],
    "U": [ ... ],
    "NC": [ ... ]
  },
  "confidence": "high" | "low",
  "edited_markdown": "FULL edited student Markdown with inline highlights"
}

Notes:
- If a code has no deductions, return an empty list for that code in "reasons".
- Keep 'evidence_snippet' short (a phrase or one sentence); if truly impossible, use an empty string.
- The edited_markdown must remain valid Markdown and keep all original content except the inline inserts for flagged parts.
"""

EXPLANATION_USER_TEMPLATE = """CONTEXT:
RUBRIC SNIPPET:
{rubric_snippet}

STUDENT MARKDOWN (original):
{student_markdown}

RAW SCORING JSON FROM PRIOR TURN:
{raw_scoring_json}

TASK:
- Explain deductions with concise reasons and evidence snippets.
- Output ONLY the JSON per the schema (no Markdown fences).
    \end{promptbox}
\caption{Prompt for the second round in which the LLM grader explains the reasons for score deductions and identifies their locations within the student's transcribed Markdown solution.}
\label{prompt: 1st round grading_system}
\end{figure*}

\begin{figure*}[t]
    \begin{promptbox}{Prompt for the LLM regrader to grade student solutions given the flagged recognition errors}
You are regrading a student's Markdown solution using the rubric and recognition-error triage. Return ONLY a single compact JSON object.

Context:
- Problem statement (possibly with illustrative images (e.g., schematics for the problem) if provided) and the student's Markdown solution (may contain some recognition errors or student's own mistakes) are provided.
- The correct final answer is provided.
- The rubric is summarized below and includes examples and max deduction caps.
- You also get the previous AI grader's outcome (per_code and deduction_reason) and the recognition-error triage (reasoning + error_details), both of which are based on the student's Markdown solution and you can locate the corresponding parts in the student's Markdown solution.

Regrade rules:
1) If recognition_errors likely caused the previous deductions (e.g., a wrong symbol/value that is inconsistent with context and appears only once),
   then IGNORE those recognition-induced issues in the second grading (do NOT deduct for them).
2) If there are student_own_mistakes in the recognition-error triage, determine (a) whether the first grading properly accounted for them or (b) whether there are also recognition errors based on the previous AI grader's outcome, the student's Markdown solution, and, the problem information. If a reasonable student mistake was NOT considered and it materially affects correctness (**be conservative when making this judgement**), deduct accordingly this time. If you think some student mistakes are just recognition errors, then do not deduct for them like 1).
3) If neither applies materially, keep the original deductions (or return empty if none).
4) Only deduct among M/E/C/U/NC. Follow the rubric_snippet; do not exceed each code's max deduction (caps will also be applied downstream).

Provide brief, concrete justification tied to the student's Markdown (and final answer), referencing recognition triage details where needed.

--- INPUTS ---
Problem statement:
{problem_statement}

Correct final answer (concise):
{final_answer}

Student's Markdown solution:
{markdown_content}

Rubric snippet (abbrev):
{rubric_snippet}

Recognition-error triage:
- reasoning: {recog_reasoning}
- error_details: {json.dumps(recog_error_details, ensure_ascii=False)[:2000]}

Previous AI grader:
- per_code: {json.dumps(orig_pred_per_code, ensure_ascii=False)}
- deduction_reason: {orig_deduction_reason}

--- OUTPUT (return only this JSON, no extra text) ---
{{
  "per_code": {{}},                // e.g., {{ "E": 0.02, "M": 0.03 }}; empty object means no deductions after regrade
  "deduction_reason": "one or two sentences explaining the deductions kept/added/removed, why, and where in the student's Markdown solution"
}}
    \end{promptbox}
\caption{Prompt used to instruct the LLM regrader (\texttt{GPT-5.1}) to evaluate student solutions while accounting for recognition errors flagged by the recognition error detector.}
\label{prompt: regrading}
\end{figure*}

\begin{figure*}[t] 
    \begin{promptbox}{Prompt for heuristically identifying recognition errors from the target recognized text (Part I)}
You are an expert assistant specializing in university-entry-level circuit analysis education and handwriting recognition analysis.
You are given a student's solution converted from handwriting to Markdown. Your task is to analyze this solution and decide, for each detected issue,
whether it is a handwriting recognition error, a student's own mistake, or uncertain.

Heuristic Principles:
- Recognition errors typically affect a local line/region; before and after that line the values/logic remain consistent.
  Student mistakes often propagate downstream and cause multiple subsequent inconsistencies.
- A single wrong line amid otherwise consistent derivation (e.g., `vb`=4V appears once while context uses 9V) suggests recognition error.
- Spurious variables that never appear in problem statement/diagram nor the student's own circuit description -> recognition_error.
- If final answer disagrees with a correct derivation at the end, triage per rules below.
- When sign or direction conventions differ but are internally consistent within the student's equations, treat them as non-errors. Only mismatched internal usage counts as error.
- Always preserve and verify all sign symbols (``-'') during recognition. Missing or faint minus signs should default to ``-'' rather than being dropped.
- When recognizing handwritten math, never reinterpret ``-'' as a hyphen or spacing artifact. If uncertain, mark it as [possible -] rather than omitting it.

Rules for Labeling Each Issue:
1) [student_own_mistake]
   When there is no contextual support to disambiguate AND the recognized value/result is completely unreasonable or far from
   the expected magnitude (e.g., should be 1.2 but shows 3457; or drastic unit/number mismatch).
   Moreover, if the ``mistake'' appears mid-derivation but the final answer matches the correct final answer, it is empirically not a student's own mistake.
   Note: Differences in current or voltage sign due solely to opposite reference directions are not student mistakes if equations are self-consistent.

2) [recognition_error]
   If context above and below is consistent and only a single line is wrong, label recognition_error.
   Also label recognition_error when a variable/value conflicts with provided diagrams or a variable never appears in the statement/diagram/description but suddenly shows up (spurious).
   Include possible OCR sign loss: if removing or adding a ``-'' restores numerical or logical consistency, label as recognition_error (missing_minus).

3) [uncertain]
   In the same ``no-context'' setting as (1), but the recognized number is close to or easily confusable with the expected value (e.g., 4 vs 9, faint minus, 1/x vs x).
   In these ambiguous cases you cannot rule out recognition vs student error, so label the issue as uncertain. Use sparingly.

4) General Consistency Checks
   - Logical/Numerical inconsistency: symbol/number changes meaning without justification (e.g., Va=5V becomes -5V/5A).
   - Malformed formulas: e.g., parallel resistors written as R1 + R2; KCL/KVL sign violations.
    \end{promptbox}
\caption{Prompt (Part I) for heuristically identifying potential recognition errors in transcriptions of student handwritten work from \dsname. The prompt includes a set of rules to help the detector exclude genuine student mistakes from flagged recognition errors. If the detector is uncertain about a particular case, it labels it as ``uncertain''. During the regrading phase, all such cases are reviewed by a human expert (the doctoral teaching assistant in our case study).}
\label{prompt: recog err detection (Part I)}
\end{figure*}

\begin{figure*}[t] 
    \begin{promptbox}{Prompt for heuristically identifying recognition errors from the target recognized text (Part II)}
--- INPUT DATA ---

1) Problem Statement:
{problem_statement}

2) Correct Final Answer (for reference):
{final_answer}

3) Recognized Student Solution (Markdown):
{markdown_content}

--- TASK ---
Read the solution and identify issues. For each issue, provide:
- a precise location_snippet from the Markdown,
- a short diagnostic,
- a tags array containing exactly one of: ["recognition_error"], ["student_own_mistake"], or ["uncertain"],
- any image indices used (e.g., "#S0", "#P1"),
- and the rule(s) applied. Also add an overall summary for human-in-the-loop routing.

--- OUTPUT FORMAT (MUST be a single valid JSON object, no extra text) ---
{{
  "issue_summary": {{
    "recognition_errors": integer,
    "student_own_mistakes": integer,
    "uncertain": integer
  }},
  "contains_recognition_error": boolean,
  "reasoning": "A concise justification of the overall decision. When applicable, cite #S*/#P* and briefly note ambiguity.",
  "error_details": [
    {{
      "location_snippet": "Exact snippet from the Markdown where the issue occurs.",
      "explanation": "Why this is an issue; reference the applied rule(s).",
      "tags": ["recognition_error" | "student_own_mistake" | "uncertain"],
      "support": {{
        "image_refs": ["#S0", "#P1"],
        "context_check": "Describe local-consistency or lack thereof (e.g., above/below lines use 9V while this line is 4V).",
        "impact_scope": "local | propagating"
      }}
    }}
  ]
}}
Return only that JSON object and nothing else.
    \end{promptbox}
\caption{Prompt (Part II) for heuristically identifying potential recognition errors in transcriptions of student handwritten work from \dsname. The \texttt{markdown\_content} field contains the MLLM-generated transcription to be analyzed, while \texttt{problem\_statement} and \texttt{final\_answer} provide question-specific context (as introduced in Appendix \ref{sec: dataset detail}) to support accurate error detection.}
\label{prompt: recog err detection (Part II)}
\end{figure*}

\end{document}